\documentclass[]{bytedance_seed}

\usepackage[toc,page,header]{appendix}

\usepackage{minitoc}

\usepackage{graphicx}
\usepackage{amsmath}

\usepackage{booktabs} 
\usepackage{colortbl}
\usepackage{amsmath}
\usepackage{amssymb}
\usepackage{mathtools}
\usepackage{amsthm}
\usepackage{wrapfig}
\definecolor{skyblue}{RGB}{203, 221, 245}
\usepackage[textsize=tiny]{todonotes}
\usepackage{multirow}
\usepackage{url}
\usepackage{xspace}
\usepackage{graphicx}
\usepackage{microtype}
\usepackage{graphicx}
\usepackage{booktabs} 
\usepackage{colortbl}
\usepackage{amsmath}
\usepackage{amssymb}
\usepackage{mathtools}
\usepackage{amsthm}
\usepackage{adjustbox} 
\usepackage{tcolorbox} 
\definecolor{skyblue}{RGB}{203, 221, 245}
\newcommand{\skyblue}{\rowcolor{skyblue}}
\usepackage{fancyhdr} 
\usepackage[numbers]{natbib}
\usepackage{multicol}

\title{ BagelVLA: Enhancing Long-Horizon Manipulation via Interleaved Vision-Language-Action Generation }

\author[*1,\dagger]{Yucheng Hu}
\author[*1,\dagger]{Jianke Zhang}
\author[*2]{Yuanfei Luo}
\author[1]{Yanjiang Guo}
\author[1]{Xiaoyu Chen}
\author[1]{Xinshu Sun}
\author[1]{Kun Feng}
\author[1]{Qingzhou Lu}
\author[2]{Sheng Chen}
\author[2]{Yangang Zhang}
\author[2,\S]{Wei Li}
\author[1,\S]{Jianyu Chen}

\affiliation[1]{Tsinghua University}
\affiliation[2]{ByteDance Seed}

\contribution[*]{Equal contribution}
\contribution[\dagger]{Joint project lead}
\contribution[\S]{Corresponding Author}

\abstract{
Equipping embodied agents with the ability to reason about tasks, foresee physical outcomes, and generate precise actions is essential for general-purpose manipulation. While recent Vision-Language-Action (VLA) models have leveraged pre-trained foundation models, they typically focus on either linguistic planning or visual forecasting in isolation. 
These methods rarely integrate both capabilities simultaneously to guide action generation, leading to suboptimal performance in complex, long-horizon manipulation tasks.
To bridge this gap, we propose BagelVLA, a unified model that integrates linguistic planning, visual forecasting, and action generation within a single framework. Initialized from a pretrained unified understanding and generative model, BagelVLA is trained to interleave textual reasoning and visual prediction directly into the action execution loop. To efficiently couple these modalities, we introduce Residual Flow Guidance (RFG)
, which initializes from current observation and leverages single-step denoising to extract predictive visual features, guiding action generation with minimal latency. Extensive experiments demonstrate that BagelVLA outperforms existing baselines by a significant margin on multiple simulated and real-world benchmarks, particularly in tasks requiring multi-stage reasoning.
}

\date{\today}
\correspondence{\email{liwei.85@bytedance.com, jianyuchen@tsinghua.edu.cn}}

\checkdata[Project Page]{\url{https://cladernyjorn.github.io/BagelVLA.github.io}}

\begin{document}
\maketitle

\begin{figure*}[h]
    \centering
    \includegraphics[width=1.0\textwidth]{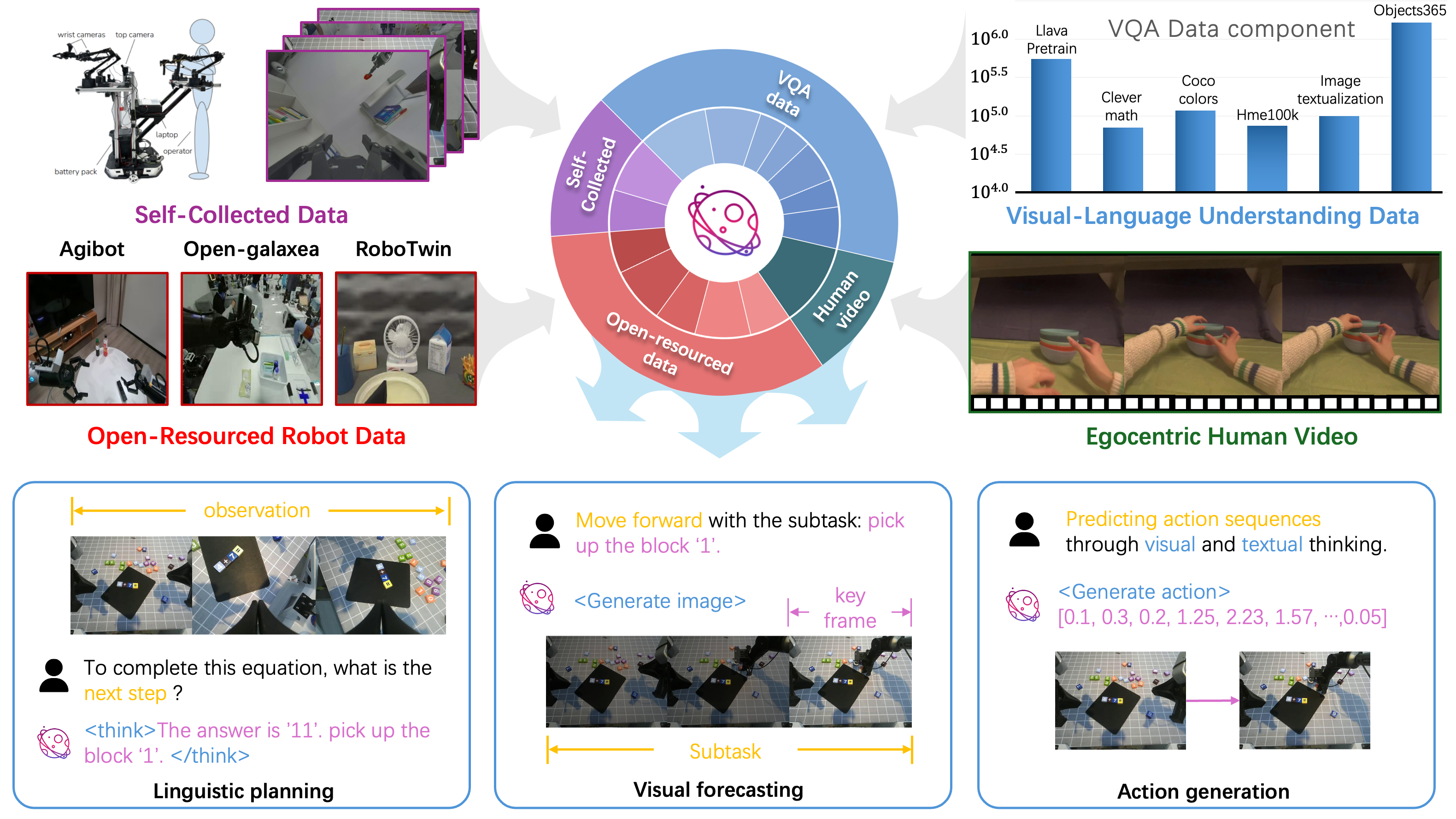}
    \caption{\textbf{Overview of our framework.} We present BagelVLA, a unified model that integrates linguistic planning, visual forecasting, and action generation within a single framework. We construct a massive hybrid dataset combining general multimodal data with large-scale robotic datasets. Robotic datasets with sub-tasks and keyframes are annotated to transfer the foundation model's general reasoning and visual generation abilities to embodied settings.}
    \label{fig:overview}
\end{figure*}
\section{Introduction}
The pursuit of generalist robots capable of performing complex manipulation tasks in unstructured environments remains a central goal in robotics. A robust embodied agent must possess three fundamental capabilities: understanding what to do based on instructions, predicting what will happen next, and executing the necessary motions. While recent Vision-Language-Action (VLA) models~\cite{brohan2023rt, team2025gr15, li2025cogvlacognitionalignedvisionlanguageactionmodel,wang2026vlingnav} have made progress by incorporating vision language models (VLMs) \citep{kim24openvla,black2024pi_0,intelligence2025pi_} or visual generation models~\citep{du2024learning,hu2024video,kim2026cosmospolicy,pai2025mimicvideo}, they often treat these capabilities as separate modules. Some methods focus on high-level planning \citep{intelligence2025pi_,fang2025robixunifiedmodelrobot} but lack visual forecasting, while others focus on visual prediction \citep{du2024learning,hu2024video,cheang2024gr2generativevideolanguageactionmodel} but struggle with the logical reasoning required for complex tasks~\cite{li2024behavior1khumancenteredembodiedai}. A unified framework that seamlessly integrates reasoning, prediction, and control remains a key challenge.

Meanwhile, the field of multimodal learning has witnessed the emergence of unified understanding and generation models~\citep{deng2025bagel,team2024chameleon,shi2024lmfusion,xie2024show,xie2025showo2improvednativeunified}. Architectures like Bagel~\cite{deng2025bagel} employ a single transformer backbone to jointly process and generate text and images, exhibiting emergent abilities in multimodal reasoning. These models provide an appealing prior for embodied agents: the model can ``think'' about the next step in text and ``imagine'' the outcome in pixels. However, such general-purpose models are not designed for embodied domain reasoning and continuous real-time control.

To make unified multimodal priors actionable for long-horizon manipulation, we propose BagelVLA, a unified VLA framework that integrates linguistic planning, visual forecasting, and action generation. Rather than treating these as isolated modules, BagelVLA interleaves them within a unified transformer architecture. The model first generates a textual plan to decompose the instruction (e.g., identifying the next object to manipulate), then predicts the future visual state, and finally generates the action. This design combines the logical reasoning of language models with the predictive power of visual generation, providing rich visual dynamics aligned with instruction to guide low-level control for long-horizon tasks.

Realizing this interleaved behavior requires a suitable training architecture and data, for which we design a two-stage training strategy to inject embodied multi-modal planning capabilities into the model.
In the first stage, we construct a massive hybrid dataset combining general multimodal data~\cite{wiedmann2025finevisionopendataneed, liu2024improvedbaselinesvisualinstruction,li2025zebracotdatasetinterleavedvision} with large-scale robotic datasets~\cite{khazatsky2024droid, agibotworldcontributors2025agibotworldcolosseolargescale, jiang2025galaxeaopenworlddatasetg0, wu2025robomindbenchmarkmultiembodimentintelligence}. Robotic datasets from diverse embodiments are annotated to transfer the model’s general reasoning and visual predictive abilities to embodied settings. In the second stage, we introduce the action expert and fine-tune the full model to couple language, predicted visual dynamics, and control. This progressive approach ensures the model retains its high-level reasoning capabilities while acquiring precise low-level control policies. 
To address the high latency in combining visual generation, we introduce Residual Flow Guidance (RFG). 
Instead of generating future frames from scratch, RFG conditions on the current observation as a strong structural prior and performs a single-step denoising to predict the residual change toward the next keyframe. This mechanism allows the model to extract predictive visual features efficiently, guiding action generation without the computational cost of full image synthesis~\citep{kim2026cosmospolicy, du2023learning}, which substantially reduces the foresight cost.

We validate BagelVLA through extensive experiments in both simulation and real-world environments. Results show that explicitly coupling linguistic planning with visual forecasting significantly improves performance over baselines, particularly in long-horizon tasks.
In real-world scenarios, BagelVLA demonstrates strong robustness, successfully generalizing to unseen instructions and diverse object arrangements where baseline methods often fail.
Our contributions are as follows:
\begin{itemize}
    \item We propose BagelVLA, which integrates linguistic planning, visual forecasting, and action generation into a single architecture. By explicitly modeling the transition from language to visual dynamics, our approach enhances reasoning and control in long-horizon tasks.
    
    By exploring various schemes for learning action representations from interleaved planning, we introduce Residual Flow Guidance (RFG), which uses the current observation as a structural prior and applies single-step denoising to capture future visual dynamics with minimal latency.
    
    \item BagelVLA substantially outperforms existing baselines in simulation benchmarks and demonstrates strong generalization to diverse instructions and environments in real-world experiments.
\end{itemize}

\section{Related Works}

\subsubsection{Vision-Language-Action Models}
Vision-Language-Action (VLA) models aim to enhance policy generalization to linguistic instructions and visual scenes by integrating vision-language models (VLMs) with action prediction. For example, methods like RT-2 \citep{brohan2023rt2} and OpenVLA \citep{kim2024openvla} employ discrete action tokens compatible with VLMs, allowing direct mapping from vision-language representations to executable actions, though this can limit expressiveness in continuous control. In contrast, approaches such as Octo \citep{octomodelteam2024octo}, 3D Diffuser Actor \citep{ke20243ddiffuseractorpolicy}, and $\pi_0$ \citep{black2024pi_0} utilize continuous action representations via diffusion models to capture multimodal distributions, better handling fine-grained manipulations. However, these methods—whether discrete or continuous, overlook the alignment gap between VLM pre-training and VLA fine-tuning, resulting in degraded vision-language capabilities during adaptation.
To mitigate this gap, other approaches \citep{zhang2025up,hu2024video,guo2024prediction,zhang2025dreamvla,kim2026cosmospolicy,pai2025mimicvideo} introduce visual prediction tasks as a bridge to map vision-language signals to action signals.
For instance, VPP \citep{hu2024video} proposes a video prediction policy that conditions robot actions on future visual representations derived from video diffusion models.
Cosmos Policy \cite{kim2026cosmospolicy} directly fine-tunes a large pretrained video model to serve as a robot policy.
Although pre-training with pixel prediction can be easily aligned with the robot observations, the absence of a dedicated VLM backbone often leads to poor instruction-following performance, particularly in tasks requiring complex reasoning.

\subsubsection{Unified understanding and generation models}
In multimodal learning, recent efforts \citep{deng2025bagel,team2024chameleon,shi2024lmfusion,xie2024show} have developed unified architectures for joint understanding and generation across modalities. For example, Bagel \citep{deng2025bagel} uses a single transformer to process and generate text and images, trained on interleaved datasets for emergent reasoning. Chameleon \citep{team2024chameleon} employs a token-based framework for mixed-modal input/output, supporting tasks like question answering and image generation. LMFusion \citep{shi2024lmfusion} integrates language and vision in a fused transformer, focusing on efficient cross-modal alignment, while Show-o \citep{xie2024show} emphasizes unified multimodal understanding and generation, including text-conditioned image generation and editing for enhanced scene comprehension. These models, trained on diverse datasets including generation, QA, and editing, demonstrate strong capabilities in multimodal reasoning that can extend to embodied agents.
Inspired by these, several VLA works \citep{lv2025f1,zhang2025unicod,chen2025villa,lu2025uniugp} have introduced action experts to transfer their capabilities to embodied scenarios. However, the lack of explicit embodied vision-language interleaved reasoning means these approaches only retain a subset of the original model’s capabilities, failing to implement step-by-step multimodal chain-of-thought reasoning. This deficiency is deemed critical for complex long-horizon tasks. In contrast, our proposed methods successfully incorporate the multi-modal reasoning capability into robotic manipulation via a complete data processing pipeline and a progressive training paradigm.

\section{Methodology}
\label{sec:method}

\subsection{Preliminaries: Interleaved planning for Robot Control}
\label{sec:problem_def}

For classic language-conditioned manipulation settings, a policy is typically learned from a demonstration dataset $\mathcal{D}=\{L_i, \tau_i\}_{i=1}^N$, where each trajectory $\tau_i=\{(v_1, l_1, a_1), \dots, (v_T, l_T, a_T)\}$ consists of observations $v_t$ (images and proprioception), stage-specific language descriptions $l_t$, and action chunks $a_t$. Conventional VLA models simplify this by conditioning purely on the global instruction $L$, learning a direct mapping policy $p_\theta(a_t|v_t, L)$. 
However, this formulation is insufficient for long-horizon tasks where a global instruction (e.g., stacking blocks in a specified order (red$\rightarrow$yellow$\rightarrow$blue$\rightarrow$green)) implicitly entails a sequence of distinct stages. We address this by modeling the problem as \textbf{Interleaved Planning}. Instead of a black-box mapping, we require the model to explicitly reason through the causal chain of the task.

Formally, given the global instruction $L$ and current observation $v_t$, BagelVLA models the joint distribution of the current subtask $l_t$, the future outcome (keyframe) $v_{t+k}$, and the action $a_t$. This joint distribution $p_\theta(a_t, v_{t+k}, l_t | v_t, L)$ is factorized based on the logical dependency of manipulation:
\begin{enumerate}
    \item Linguistic Planning: The model first identifies the immediate textual objective $l_t$ from the global instruction. We consider task decomposition to be the primary semantic capability of VLM-based architectures.
    \item Visual Forecasting: Conditioned on this subtask, the model acts as a world model to predict the physical outcome $v_{t+k}$.
    \item Action Generation: Finally, the action $a_t$ is generated, grounded in both the textual plan and the visual forecast.
\end{enumerate}

Consequently, our objective is formulated as the maximization of the following factorized likelihood:
\begin{align}
\notag
        \mathcal{J}&=-(\mathcal{L}_{l}+\mathcal{L}_{v}+\mathcal{L}_{a})\\
        \notag
        &=\max_\theta \mathbb{E}_{\mathcal{D}}  \log ~p_\theta(l_t|v_t, L) \cdot p_\theta(v_{t+k}|v_t, L, l_t)
    \cdot p_\theta(a_t|v_t, L, l_t, v_{t+k}) 
\end{align}

\begin{figure*}[ht]
    \centering
    \includegraphics[width=0.96\textwidth]{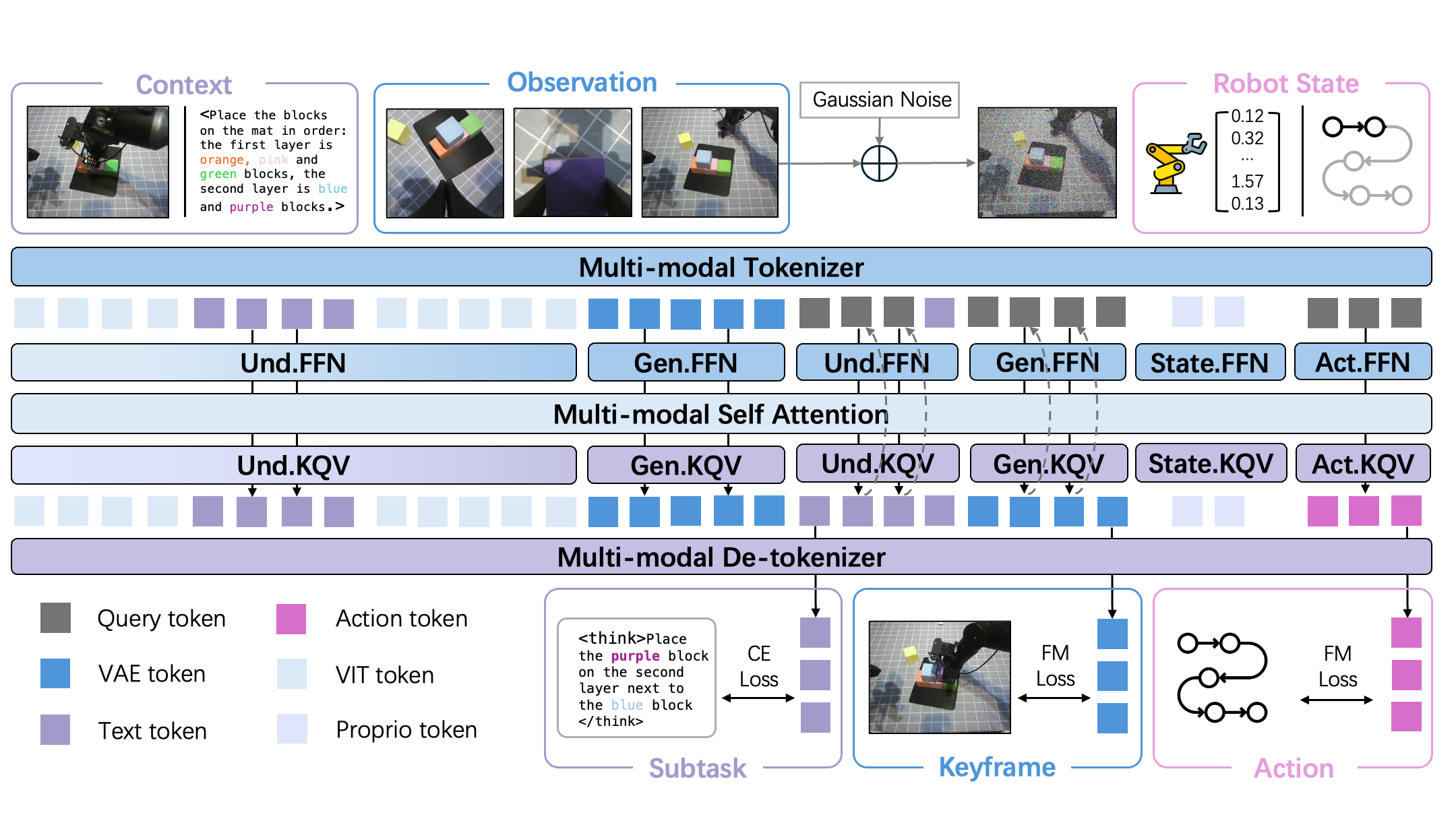}
    \caption{\textbf{Illustration of the BagelVLA framework.}  BagelVLA~ utilizes a Mixture-of-Transformers (MoT) architecture, comprising three independent transformers specialized for linguistic, visual, and action modalities. To tackle long-horizon tasks and semantic generalization, we formulate language-conditioned action learning as a long-sequence interleaved planning problem. As shown, we structure these modalities into a unified sequence, enabling the model to generate predictions across all three modalities based on the interleaved context. To support this architecture, we have designed specific mechanisms to facilitate interaction among multiple flow-matching experts and to enhance inference efficiency.}
    \label{fig:main_chart}
\end{figure*}

\subsection{Model Architecture}
To address the interleaved planning problem defined in Sec. \ref{sec:problem_def}, we propose \textbf{BagelVLA}, a unified framework for understanding, prediction, and action generation. As illustrated in Fig. \ref{fig:main_chart}, BagelVLA is designed to process data across three modalities simultaneously. To leverage pre-existing large-scale multimodal data, we employ a Mixture of Transformers (MoT) architecture to orchestrate experts managing different modalities: specifically, an LLM expert, a generation expert, and an action expert, all connected via self-attention mechanisms.

We initialize the LLM and generation experts using Bagel \cite{deng2025bagel}, a unified MoT model for understanding and generation. On top of this foundation, we incorporate a smaller transformer to serve as the action expert. Distinct from prior MoT-based VLA architectures \citep{lv2025f1,zhang2025unicod,chen2025villa}, BagelVLA benefits from robust pre-training initialization for both its language and vision components and employs a novel dual flow-matching mechanism (detailed in Sec. \ref{sec:dualfm}). Detailed model settings are described in Appendix \ref{sec:app-model}.

\subsubsection{Understanding Expert \& Generation Expert}

The understanding and generation experts 
adopt the architecture of Qwen2.5-LLM-7B \cite{qwen2.5}. Following Bagel's configuration, we utilize two distinct visual encoders responsible for visual-language understanding and goal image prediction, respectively.
Each input observation view $v_t$ is encoded by the SigLIP2 \cite{tschannen2025siglip} and concatenated with the text instructions $L$ (and $l_t$) to serve as input for the LLM Expert. 

We also utilize the VAE from FLUX \cite{flux2024} to encode images. 
For \textbf{linguistic planning}, the understanding expert attends to ViT features when autoregressively generating the subtask $l_t$. We optimize textual-planning task using an autoregressive Cross-Entropy (CE) loss: $\mathcal{L}_{l} = -\log p_\theta(l_t|v_t, L)$. 
For \textbf{visual forecasting}, the generation expert, while denoising the keyframe image, attends to all input views' VAE and ViT features, and relevant textual information. It generates keyframe by iteratively denoising input VAE noise using Flow Matching~\citep{lipman2023fm,liu2023rflow}, denoted as: $\mathcal{L}_{v}=-\log p_\theta(v_{t+k}|v_t, L, l_t)$.

\subsubsection{Action Expert}

We employ an independent transformer connected via the MoT framework as the action expert, which is responsible for processing proprioceptive and action modalities. The action expert shares a similar architecture with the Qwen2.5 LLM; however, we reduce the intermediate size of the MLP to 1/5th of the original, resulting in 2B parameters. This compact size facilitates higher execution frequency during inference through KV-cache and asynchronous action generation.

For \textbf{action planning}, we employ Flow Matching to learn action chunks, denoted as $\mathcal{L}_{a}=-\log p_\theta(a_t|v_t, L, l_t, v_{t+k}) $. During the denoising process, the action sequence can attend to the VAE and ViT features of the input views, the global instruction $L$, the generated subtask $l_t$, and also the proprioceptive state input to the action expert. 
Notably, the action expert attends to the intermediate latent states of the image currently being generated. 
This involves handling the asymmetric information interaction between the dual Flow Matching modules of the generation and action experts. We detail the various schemes we explored in Sec. \ref{sec:dualfm} and ablate these methods in Sec. \ref{sec:ablation}.

\subsection{Conditioning Schemes in Dual Flow-Matching}
\label{sec:dualfm}
\begin{figure}[ht]
    \centering
    \includegraphics[width=0.8\textwidth]{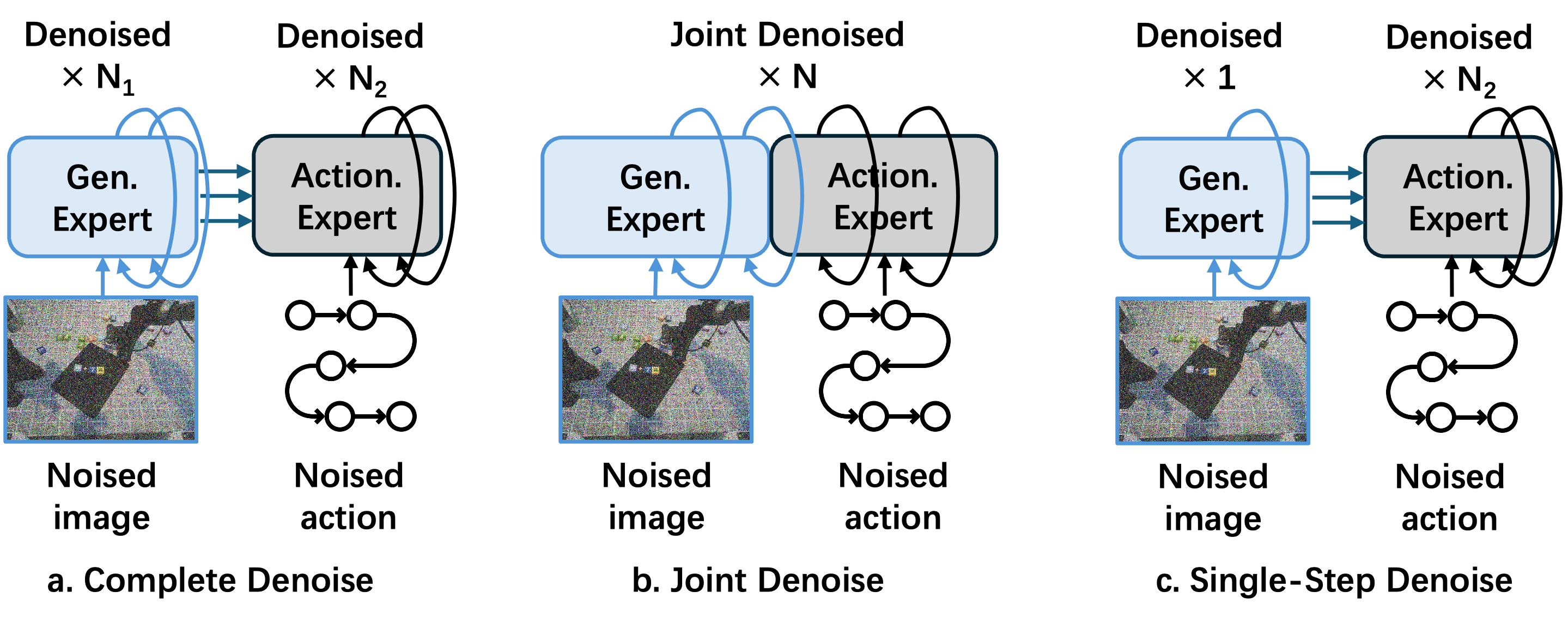}
    \caption{\textbf{Illustration of different types of dual denoising schemes.} (a) Complete Denoise: Image prediction and action generation are performed separately, requiring a total of $N_1+N_2$ denoising steps.
(b) Joint Denoise: Image prediction and action generation are performed simultaneously, denoising together for $N$ steps.
(c) Single-Step Denoise: Action generation is conditioned directly on the context from the first denoising step of the image prediction. Further implementation details, including the construction of the concatenated sequence and the attention mask are provided in Appendix \ref{sec:app-dualfm}.
} 
    \label{fig:interaction}
\end{figure}
In this section, we detail the computation of $\mathcal{L}_v$ and $\mathcal{L}_a$ within a unified interleaved input sequence, ensuring consistency with the inference context. As illustrated in Fig. \ref{fig:interaction}, we propose three interaction mechanisms for the Flow Matching (FM) of keyframe prediction and action generation.

\paragraph{Scheme 1: \textbf{Complete Denoise}}
As shown in Fig. \ref{fig:interaction}(a), Complete Denoise prioritizes the full denoising of the keyframe image by the generation expert. The generated image is then fed back as context for action generation. During training, to ensure the action expert observes a fully denoised image, we append the ground truth keyframe subsequent to the denoising sequence. The loss functions are defined as follows:
\begin{align}
\notag
    \mathcal{L}_v = \mathbb{E} &\left[ || \mathbf{v}_{v,\theta}(L, v_t, l_t, \tau, v_{t+k}^\tau) - (v_{t+k}^1 - v_{t+k}^0) ||_2^2 \right], ~~v_{t+k}^\tau = (1-\tau)v_{t+k}^0 + \tau v_{t+k}^1, ~~v_{t+k}^1=v_{t+k}\\
    \label{eq-lv}
    \mathcal{L}_{a1} = \mathbb{E} &\left[ || \mathbf{v}_{a,\theta}(L, v_t, l_t, v_{t+k}^{\tau=1}, \tau, a_t^\tau) - (a_t^1 - a_t^0) ||_2^2 \right], ~~~a_t^\tau = (1-\tau)a_t^0 + \tau a_t^1, ~~a_t^1=a_t
\end{align}
where $\mathbf{v}$ denotes the velocity predicted by the model for the corresponding modality. $L$ and $l_t$ represent the global instruction and sub-task plan, $v_t$ is the current input observation, $v_{t+k}$ is the target keyframe and $a_t$ is the action chunk. $\tau$ denotes the denoising timestep (where $\tau=0$ represents initial noise and $\tau=1$ represents the ground truth target).

This approach effectively combines a World Model (WM) with an Inverse Dynamics Model (IDM)~\citep{du2024learning}. While theoretically sound for leveraging the WM, it suffers from high inference latency (total denoising steps $N_1 + N_2$) and the potential accumulation of visual errors. To mitigate these issues, we propose alternative schemes.

\paragraph{Scheme 2: \textbf{Joint Denoise}}
As shown in Fig. \ref{fig:interaction}(b), we synchronize the denoising processes of the keyframe and the action. Here, the action generation attends to the noisy image currently undergoing denoising. The computation for the action FM loss in Eq. \ref{eq-lv} is modified as:
\begin{equation}
\notag
    \mathcal{L}_{a2} = \mathbb{E} \left[ || \mathbf{v}_{a,\theta}(L, v_t, l_t, v_{t+k}^\tau, \tau, a_t^\tau) - (a_t^1 - a_t^0) ||_2^2 \right]
\end{equation}
During training, we append the action denoising block directly after the keyframe denoising sequence, allowing the action component to attend to the intermediate noisy keyframes. During inference, the model generates both keyframes and actions within $N$ steps, significantly reducing latency.

\paragraph{Scheme 3: \textbf{Single-step Denoise}}
To further minimize the computational cost of action inference imposed by image denoising, we propose single-step denoise. In this scheme, action generation attends only to the KV-cache from the initial denoising step of the keyframe. This implies the model generates actions while conditioning on the initial noise as the keyframe input:
\begin{equation}
\notag
    \mathcal{L}_{a3} = \mathbb{E} \left[ || \mathbf{v}_{a,\theta}(L, v_t, l_t, v_{t+k}^{\tau=0}, \tau, a_t^\tau) - (a_t^1 - a_t^0) ||_2^2 \right]
\end{equation}

Furthermore, based on Scheme 3, we introduce a variant of Single-step Denoise where we inject current frame information into the initial image noise to provide stronger priors for both keyframe and action generation:
\begin{align}
\label{vpp1}
    &\text{Naive Single-step Denoise}: ~\qquad v_{t+k}^{\tau=0} \sim \mathcal{N}(0, I)\\
\label{vpp2}
    &\text{Residual Flow Guidance } \textbf{(RFG)}: v_{t+k}^{\tau=0} \sim \mathcal{N}(v_t, I)
\end{align}
More details about implementing the above methods can be found in Appendix \ref{sec:app-dualfm}. 
We provide an ablation study of these methods in Sec. \ref{sec:ablation}. Based on the results, we select the Single-step Denoise (RFG) as our default setting for BagelVLA. Notably, we observe that RFG, which incorporates the initial frame $v_t$ prior, significantly reduces the required denoising steps as shown in Fig. \ref{fig:keyframe_prediction}. We hypothesize that this allows the WM to focus on modeling robot manipulation changes rather than reconstructing static background details. Further quantitative comparisons are available in Sec. \ref{sec:exp-dualfm}.

\subsection{Data Engine}
To construct a large-scale pretraining dataset for subtask planning and keyframe prediction in embodied scenarios, we leverage diverse sources of manipulation demonstrations and apply tailored processing pipelines to four major data categories in Fig. \ref{fig:overview} according to their characteristics. Details of all data annotations and components are provided in Appendix \ref{sec:app-data}.

\begin{itemize}
    \item \textbf{Robotic Data} The robot data comprises self-collected expert demonstrations and publicly available data from diverse embodiments. For proprietary data, we manually annotate and segment videos to obtain $l_t$, ensuring high-quality planning and keyframe prediction. For public datasets lacking fine-grained labels,  we utilize Seed-1.5-VL-thinking \cite{guo2025seed15vltechnicalreport} to synthesize $l_t$ and identify temporal boundaries (start and end frames). These samples are then filtered to retain high-quality instances. These two components are used exclusively for pretraining to transfer the model’s fundamental planning and prediction capabilities to the embodied domain.
    \item \textbf{General Data} General Data includes egocentric human videos and large-scale image–text VQA data. For the former, we similarly employ Seed-1.5-VL-thinking to generate language annotations; however, due to the complexity of human-centered scenes, we do not annotate subtasks and instead predict only the final frame of each operation.
\end{itemize}
These two data sources are mainly used to preserve the base model’s original understanding and generation capabilities. 

\subsection{Training and Inference Strategy}
BagelVLA requires the simultaneous alignment of three distinct planning tasks: linguistic planning, visual forecasting, and action generation. To achieve this, we divide the training process into two stages, maximizing the utilization of general multimodal data and large-scale embodied data without action labels.
Detailed data recipes and implementation details can be found in Appendix \ref{sec:app-data} and \ref{sec:app-impl}.

\paragraph{Stage 1: Pretraining - Finetuning Linguistic Planning and Learning Visual Dynamics}
In this stage, we exclusively finetune the understanding and generation experts to acquire capabilities in textual planning and keyframe prediction. To preserve the model's general linguistic proficiency, we co-train with general Question-Answering (QA) data.
Specifically, the pretraining data comprises:
\begin{itemize}
    \item General VQA (Language Co-training): 2.98M QA pairs.
    \item Human-hand Data (Visual Dynamics): 310k episodes.
    \item Open-source Robot Data (Language Planning \& Visual Dynamics): 146k episodes.
    \item Open-source Robot Data (Visual Dynamics): 297k episodes.
    \item Self-collected Real Robot Data (Language Planning \& Visual Dynamics): 75k episodes.
\end{itemize}

\paragraph{Stage 2: Finetuning - Learning Action Planning}
In this stage, we introduce downstream robot data containing action labels for finetuning. We finetune the entire model on all three planning tasks simultaneously to obtain an interleaved planning model that performs robustly in specific scenarios. 
For the four scenarios used in our experiments, we employ the following finetuning strategies:
\begin{itemize}
    \item Calvin (Visual \& Action Planning): ABC split dataset.
    \item Robotwin (Linguistic, Visual \& Action Planning): 50 tasks with 50 episodes each, totaling 2.5k episodes.
    \item ALOHA Basic Tasks (Visual \& Action Planning): 3k episodes.
    \item ALOHA Long-horizon Tasks (Linguistic, Visual \& Action Planning): 1.5k episodes.
\end{itemize}

\textbf{Inference Strategy}
During inference, the model generates textual plans, keyframes, and actions in an interleaved manner.
At each denoising step, only a single expert is activated (7B model for text and keyframe or 2B model for action generation).
The single-step denoise scheme further enhances execution frequency. Specifically, we concatenate the current frame, instruction context, and a pure noise image to compute the KV pairs of the understanding and generation experts, which then condition the action generation. This mechanism enables BagelVLA to infer at a speed of 1.2 seconds per chunk on a single RTX 5090 GPU (yielding a real-world action frequency of 40Hz with a chunk size of 48).

We also introduce Asynchronous Execution \cite{zhang2024hirt,cui2025openhelix} to further boost inference speed. During training, we randomly replace the current frame with a preceding image. This allows us to reduce the updating frequency of the KV contexts of understanding and generation experts during inference, updating only the proprioceptive inputs to output new action chunks. Under this setting, our policy can achieve an execution frequency of 72Hz.

\section{Experiment}
We conduct extensive experiments to evaluate the interleaved planning capabilities of BagelVLA across a diverse range of manipulation tasks. These experiments encompass two simulation environments, Calvin \cite{mees2022calvin} and Robotwin \cite{chen2025robotwin}, as well as a basic tasks suite containing 9 skills of 30 tasks, and a long-horizon task suite performed on the AgileX dual-arm robot system.

\subsection{Evaluation in Simulation Environment}
We benchmark BagelVLA against $\pi_0$\cite{black2024pi_0}, RDT \cite{liu2024rdt} and two VLA models that incorporate future prediction capabilities, UP-VLA \cite{zhang2025up} and VPP \cite{hu2024video}, in the Calvin and Robotwin environments.

In the Calvin environment, models are trained on the ABC split and evaluated in the D environment. For Robotwin, we utilize a training dataset consisting of 50 clean demonstrations for each of the 50 tasks. All models are then tested in both Clean and Randomized settings using unseen instructions. To verify the efficacy of interleaved planning, we conduct experiments with BagelVLA trained and tested both with and without interleaved planning. 
Further details regarding simulation experiments can be found in Appendix \ref{sec:app-sim_details}.
\begin{table}[h]
\caption{\textbf{Results on Calvin and Robotwin2.0 Benchmarks.} Since Calvin consists exclusively of single-step tasks, we did not evaluate BagelVLA's performance under the textual planning setting in this domain.
Detailed results can be found in Table \ref{tab:app-calvin} and \ref{tab:app-robotwin}.
} 
\label{tab:sim_results}      
\centering
\scriptsize
\begin{adjustbox}{width=0.5\textwidth}
\begin{tabular}{l|c|cc}
\toprule
\multirow{2}{*}[-0.5ex]{\textbf{Model}}&\multicolumn{1}{c}{Calvin} & \multicolumn{2}{c}{Robotwin}\\
\cmidrule(lr){2-4} &ABC-D& Clean & Randomized\\
\midrule
$\pi_0$& 3.648&46.42 & 16.34\\
RDT &-&34.50&13.72\\
UP-VLA & 4.078&52.92 & 15.16\\
VPP & 4.329&- & -\\
w/o Textual-planning& - & 54.00 & 19.20 \\
w/o Keyframe-forecasting& 3.345 &  56.72 &  15.92 \\
\skyblue BagelVLA & \textbf{4.405} & \textbf{75.26} & \textbf{20.87} \\
\bottomrule
\end{tabular}
\end{adjustbox}
\end{table}



As presented in Table \ref{tab:sim_results}, BagelVLA outperforms all baselines on both the Calvin ABC-D split and the Robotwin tasks.
BagelVLA achieves an average completion length of 4.41 on the Calvin ABC-D benchmark. This indicates that models leveraging only visual prediction as an auxiliary task can effectively generalize from in-domain training to Out-of-Distribution (OOD) scenarios involving background and color variations, while maintaining high manipulation accuracy.

On the Robotwin benchmark, BagelVLA without textual-planning surpasses $\pi_0$ in both Clean and Randomized settings, achieving success rates comparable to UP-VLA, which similarly employs visual prediction as an auxiliary task. This suggests that the visual prediction component within our interleaved planning framework yields consistent gains across different VLM backbones. 
However, when incorporating textual-planning, BagelVLA achieves state-of-the-art performance in both in-domain and out-of-domain settings on Robotwin, demonstrating the substantial effectiveness of the proposed interleaved planning scheme.

\subsection{Real-world Experiments}
\begin{figure}[t]
    \centering
    \includegraphics[width=1.0\textwidth]{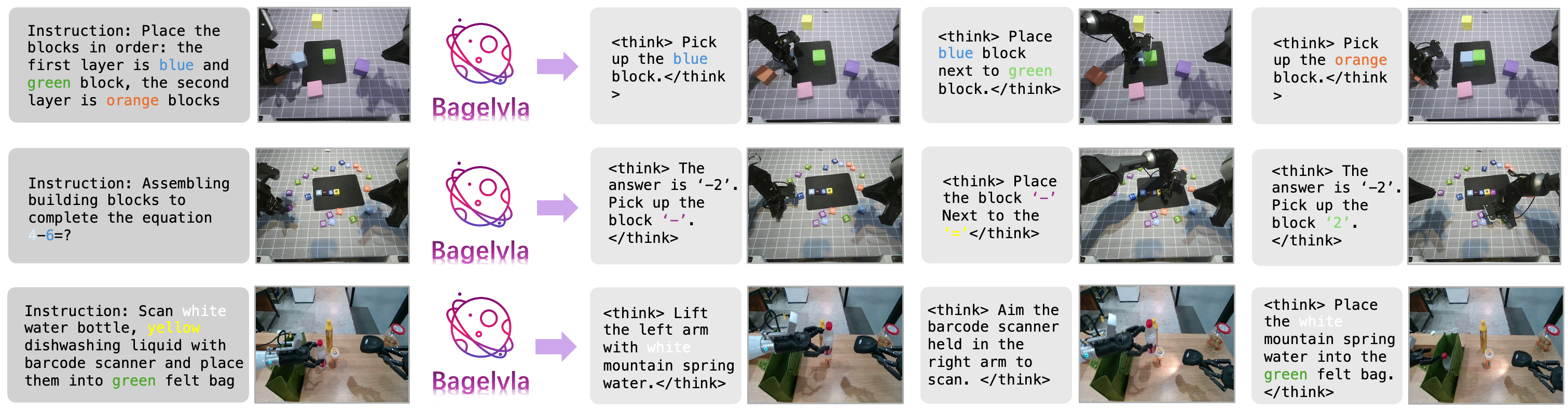}
    \caption{\textbf{Visualization of interleaved planning results on real-world robotic tasks. 
    } Given a global instruction and the current observation, BagelVLA leverages the context to identify the immediate subtask, predicts a goal image for that subtask, and subsequently generates actions. The figure illustrates the interleaved planning results for \textit{Stack Cubes in Requested Order}, \textit{Calculate and Place Symbol Blocks} task, and a task from the Agibot dataset~\citep{agibotworldcontributors2025agibotworldcolosseolargescale}. More cases can be found in Appendix \ref{sec:app-inter}.}
    \label{fig:exp-realwd}
\end{figure}
We evaluated BagelVLA on the Aloha-AgileX bimanual robot platform across two categories of dual-arm manipulation tasks. Multiple demonstrations of real-world evaluation are presented in Appendix \ref{sec:app-demo} for reference. These tasks were designed to assess the model's performance on both basic tasks and long-horizon tasks that require planning. Specifically, we collect 3,000 trajectories categorized as basic tasks, covering 9 distinct skills ranging from short-horizon tasks such as pick-and-place to medium-horizon tasks such as sweeping rubbish. Furthermore, we designed two types of Long-Horizon planning tasks that necessitate subtask planning, for which we gathered 1,500 demonstrations.
All collected data are manually annotated with subtasks and corresponding keyframes. We then fine-tune the pretrained BagelVLA on all trajectories and evaluate its multi-task learning capabilities.
We compare BagelVLA with $\pi_0$ \cite{black2024pi_0} and VPP \cite{hu2024video}. A visualization of interleaved plans generated by BagelVLA for the real-world tasks is illustrated in Fig. \ref{fig:exp-realwd}.

\subsubsection{Basic Task Experiments}
\begin{table*}[h]
\caption{\textbf{Results on Real-World Basic Tasks} without using subtask planning. We run 20 times for each task. More details and evaluation demos can be found in Appendix \ref{sec:app-basic}} 
\label{tab:short_horizon_tasks}      
\centering
\scriptsize
\begin{adjustbox}{width=\textwidth}
\begin{tabular}{r|ccccccccccc}
\toprule
\multirow{2}{*}[-0.5ex]{\textbf{Model}}& \multicolumn{1}{c}{\textbf{Pick\&Place}} & \multicolumn{1}{c}{\textbf{Pick\&Place}}&\multicolumn{1}{c}{\textbf{Water}} & \multicolumn{1}{c}{\textbf{Stack}}& \multicolumn{1}{c}{\textbf{Put Flowers}}& \multicolumn{1}{c}{\textbf{Stack}}& \multicolumn{1}{c}{\textbf{Pour}}& \multicolumn{1}{c}{\textbf{Sweep}}& \multicolumn{1}{c}{\textbf{Press}}&\multicolumn{1}{c}{\textbf{Drawer}}&\multicolumn{1}{c}{\textbf{Success}} \\
 &\textbf{Seen} & \textbf{Unseen} & \textbf{Flower} & \textbf{Cubes}& \textbf{in Vase} & \textbf{Bowls} & \textbf{Fries} & \textbf{Rubbish} & \textbf{Button}& \textbf{Close} & \textbf{Average} \\
\midrule
$\pi_0$ & \textbf{95} & 55 & 50 & 65 & 40 & 70 & 35 & 55 & \textbf{90} & 95 & 65.0 \\
VPP & 85 & 45 & \textbf{60} & 50 & \textbf{50} & 55 & 30 & 45 & 75 & 100 & 59.5 \\
\skyblue BagelVLA & \textbf{95} & \textbf{85} & \textbf{60} & \textbf{80} & 35 & \textbf{90} & \textbf{45} & \textbf{80} & \textbf{90} & 95 & \textbf{75.5} \\
\bottomrule
\end{tabular}
\end{adjustbox}
\end{table*}
Table~\ref{tab:short_horizon_tasks} presents the performance of BagelVLA in a multi-task setting without the use of subtask planning. BagelVLA achieved the highest average success rate across the 9 categories of tasks, which demonstrates its outstanding multi-task learning capabilities. Additionally, we tested the model on pick-and-place tasks involving unseen objects. As shown, BagelVLA significantly outperforms VPP and $\pi_0$ in the OOD setting. This advantage stems from the powerful semantic features preserved during VLA fine-tuning, which are inherited from the pre-training of our understanding and generation experts. 
\subsubsection{Long-Horizon Planning Task Experiments}
\label{sec:exp-hard}
\begin{table*}[h]
\caption{\textbf{Results on Real-World Long-Horizon Planning Tasks.} We run 20 times for each task. Difficulty settings for the tasks are defined as follows. For Stack Cubes: Easy (2-3 cubes, 1-2 layers), Middle (3-4 cubes, 2 layers), and Hard (3-5 cubes, 3 layers). For Calculate and Place Symbols: Easy (1-2 blocks for single-digit addition), Middle (2 blocks for the answer of a double-digit addition), and Hard (3-5 blocks within a double-digit addition). The Success Rate column shows the average success rate, while \textit{Planning Accuracy} indicates correct motion trends across 20 different intermediate states. More details and evaluation demos can be found in Appendix \ref{sec:app-hard}.
}
\label{tab:long_horizon_tasks}
\centering
\scriptsize
\begin{adjustbox}{width=\textwidth}
\begin{tabular}{r|ccc|cc|ccc|cc}
\toprule
\multicolumn{1}{c|}{\textbf{Tasks}} & \multicolumn{5}{c|}{\textbf{Stack Cubes in Requested Order}} & \multicolumn{5}{c}{\textbf{Calculate and Place Symbol Blocks}} \\
\cmidrule(lr){1-6} \cmidrule(lr){7-11}
\textbf{Difficulty} & 
{Easy} & 
{Middle} & 
{Hard } & 
{Success Rate$\uparrow$} & 
{Planning Accuracy$\uparrow$} & 
{Easy} & 
{Middle} & 
{Hard} & 
{Success Rate$\uparrow$} & 
{Planning Accuracy$\uparrow$}\\
\midrule
$\pi_0$ & 75 & 35 & 10 & 40.0 & 55 & 70 & 25 & 0 & 31.7 & 40 \\
VPP & 60 & 15 &  0 & 25.0 &  45 & 60 & 10 &  0 & 23.3 & 30 \\
w/o Keyframe-forecasting & 90 & 45 & 25 & 53.3 & 80 & 70 & 50 & 30 & 50.0 & 75 \\
w/o Textual-planning & 75 & 40 & 15 & 43.3 & 70 & 65 & 25 & 10 & 33.3 & 50 \\
\skyblue BagelVLA & \textbf{95} &  \textbf{65} & \textbf{60} & \textbf{73.3} & \textbf{95} &  \textbf{80} &  \textbf{65} & \textbf{45} &\textbf{63.3} & \textbf{85} \\
\bottomrule
\end{tabular}
\end{adjustbox}
\end{table*}
We collected data for two categories of long-horizon tasks that require planning. In the colored block stacking task, shown in the first row of Fig. \ref{fig:exp-realwd}, the model is instructed to stack cubes in the order specified by the instruction. This task challenges both the model's visual-language interleave planning ability and its capacity to follow instructions at the action level. In the arithmetic equation arrangement task, shown in the second row, we require the model first to compute an arithmetic expression and then place the corresponding symbolic blocks in a single sequence. The objective of this task is to verify whether the model can retain reasoning capabilities (such as performing simple addition) during the planning process.
Table~\ref{tab:long_horizon_tasks} displays the performance of the three models on these two long-horizon planning tasks. It is evident that although all three models were trained on the exact same action data, BagelVLA, with its interleaved planning capabilities, exhibits a significant advantage in planning-oriented tasks. In addition to the average task success rate, we also measured the correctness of the motion trend for each subtask to assess the model's semantic understanding and action-following fidelity. Overall, BagelVLA achieved a planning accuracy of nearly 90\%, which implies that its multi-modal planning is correct and possesses strong generalization abilities. Concurrently, we observed a gap between task success rate and the planning accuracy, suggesting deficiencies in action mapping due to limitations in both the model and the dataset, specifically concerning the precision of fine-motor control.
\subsection{Ablation Study}
\label{sec:ablation}
We conduct comprehensive ablation studies on the various modules of BagelVLA in both simulated and real-world environments. Through these experiments, we aim to answer the following questions:
\begin{enumerate}
    \item What is the optimal interaction mechanism between the generation experts and the action experts?
    \item How does RFG outperform naive single-step denoising?
    \item What is the effect of BagelVLA's pre-training on action execution performance?
    \item Does each modality within the interleaved planning framework contribute positively to the action generation process?
\end{enumerate}
\subsubsection{\textbf{Comparison of \textbf{Different Conditioning Schemes} in Dual Flow-Matching}}
\label{sec:exp-dualfm}
We evaluate the three dual flow-matching interaction schemes described in Sec.~\ref{sec:dualfm} within the Calvin ABC-D environment. For complete denoise and joint denoise, the image noise initialization follows the formulation in Eq.~\ref{vpp1}. We utilize single-view inputs and train each method for 10k steps for testing. 
We denoise 50 times for image generation and 10 times for action generation. Table~\ref{tab:interaction_comparison} reports the average task completion length and the inference latency per action chunk for each interaction method, measured 20 times on a single NVIDIA A800 GPU.
\begin{table}[h]
\caption{Ablation on Different Conditioning Schemes and RFG mechanism on Calvin single-view setting.
} 
\label{tab:interaction_comparison}      
\centering
\scriptsize
\begin{adjustbox}{width=0.55\textwidth}
\begin{tabular}{l|c|cc}
\toprule
\multicolumn{1}{c}{\textbf{Dual Flow-Matching Schemes}}&\multicolumn{1}{c}{\textbf{Latency$\downarrow$}} & \multicolumn{1}{c}{\textbf{ABC-D$\uparrow$}}\\
\midrule
Complete Denoise & 6.04s & 2.480 \\
Joint Denoise & 2.90s & 2.038 \\
Single-step Denoise (Eq.~\ref{vpp1}) & \textbf{1.23s} & 3.345 \\
RFG (Eq.~\ref{vpp2})& \textbf{1.23s} & \textbf{3.600} \\
\bottomrule
\end{tabular}
\end{adjustbox}
\vspace{-2ex}
\end{table}

The results indicate that the \textit{single-step denoising} strategy not only significantly outperforms the other two approaches in terms of task success rate but also achieves superior inference speed. This performance gap can be attributed to the domain shift introduced during testing, where the model encounters scenes with altered color schemes. Under these conditions, models employing \textit{complete denoising} or \textit{joint denoising} are prone to encountering out-of-distribution (OOD) intermediate states during the flow-matching phase of the generation expert. This consequently leads to a substantial degradation in action execution performance. Based on these empirical findings, we adopt \textit{single-step denoising} as the default interaction mechanism for the dual flow-matching framework in BagelVLA across all subsequent scenarios and tasks.

\subsubsection{\textbf{Advantages of \textbf{RFG over Naive Single-Step Denoising}}}
\begin{figure*}[ht]
    \centering
    \includegraphics[width=0.98\textwidth]{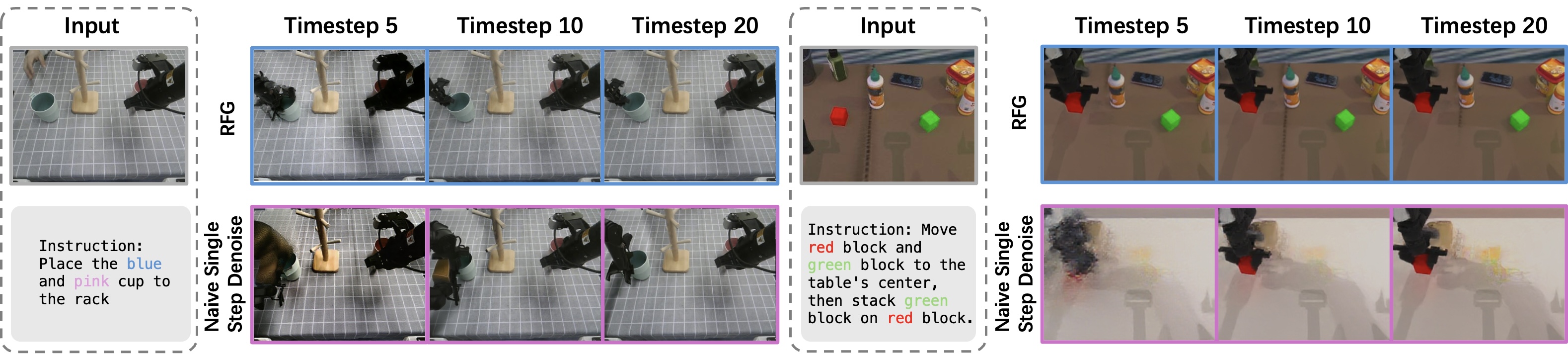}
    \caption{\textbf{Predicted images using different denoising steps. 
    } The figure displays the generation results for the naive single-step denoise (Eq.~\ref{vpp1}) and RFG (Eq.~\ref{vpp2}) across varying denoising steps in real-world basic tasks and Robotwin randomized (unseen) scenarios. RFG demonstrates the capability to preserve backgrounds and achieve high-quality generation with very few steps. This provides strong support for reducing the inference latency of interleaved generation. More cases can be found in Appendix \ref{sec:app-rfg}.}
    \label{fig:keyframe_prediction}
\end{figure*}
\begin{figure}[ht]
    \centering
    \includegraphics[width=0.55\textwidth]{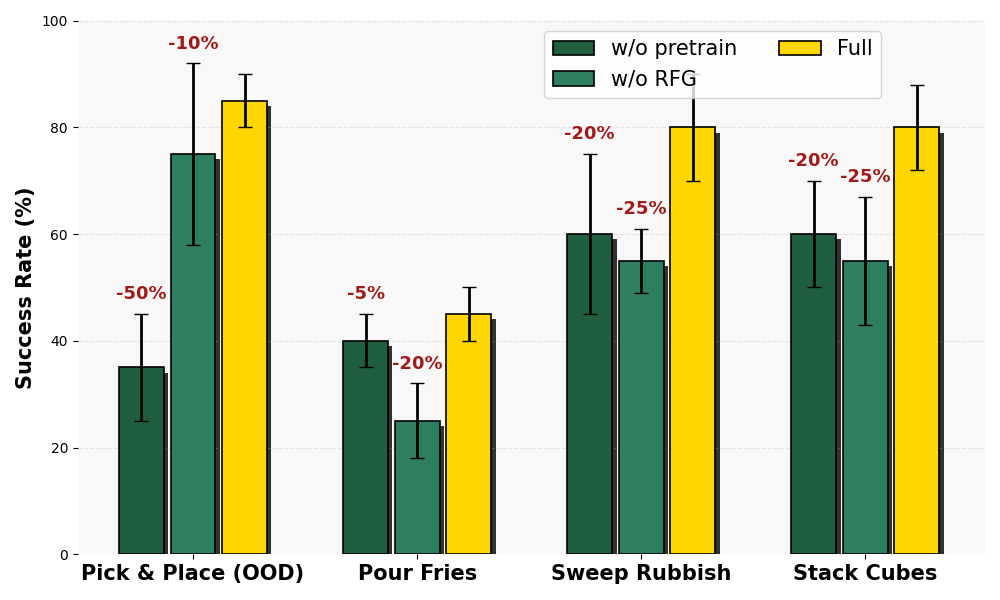}
    \caption{\textbf{Ablation on Real-World Basic Tasks}}
    \label{fig:abl-norm}
\end{figure}
In contrast to the conventional naive single-step denoising approach, which employs Eq.~\ref{vpp1} for noise initialization, RFG utilizes Eq.~\ref{vpp2}. We compare these two methods across both the Calvin simulation environment and real-world basic tasks. In the real-world basic tasks shown in Fig.~\ref{fig:abl-norm}, RFG demonstrates significantly superior performance compared to naive single-step denoising on several tasks. Concurrently, 
 as shown in Table~\ref{tab:interaction_comparison}, RFG achieves faster action learning convergence while maintaining the low inference latency characteristic of naive single-step denoising. This improvement stems from the fact that in the naive approach, action generation relies on intermediate features derived from a single denoising step on pure Gaussian noise. Conversely, RFG incorporates the initial frame into the noise initialization, thereby providing stronger prior information for action generation.

Furthermore, we observe that RFG offers a distinct advantage in keyframe prediction, even though fully denoising the keyframe is not strictly required for action generation. Fig.~\ref{fig:keyframe_prediction} visualizes the predicted keyframes for both RFG and naive single-step denoising across different denoising steps. It is evident that RFG is capable of generating high-quality future frames with very few denoising steps (e.g., 10 steps). We hypothesize that this phenomenon arises because the inclusion of the first frame in Eq.~\ref{vpp2} allows the model to focus its capacity on the dynamic regions, rather than learning complex static background information.

\subsubsection{\textbf{Effectiveness of Large-Scale Language Planning and Visual Dynamics \textbf{Pre-training}}}
In the real-world basic tasks, we evaluate the impact of pre-training. By comparing the \texttt{w/o pretrain} variant with \texttt{baseline} in Fig.~\ref{fig:abl-norm}, it is evident that the pre-trained baseline achieves a significantly higher success rate on \texttt{pick\&place (OOD)}tasks. This indicates that pre-training solely on linguistic planning and visual forecasting is sufficient to enhance the model's semantic generalization capabilities. 
Furthermore, on three medium-horizon tasks (including \texttt{sweep rubbish}, \texttt{pour fries} and \texttt{stack cubes}), the model utilizing joint pre-training exhibits higher accuracy. We attribute this improvement to the planning capabilities acquired from the language planning tasks during pre-training. During the subsequent action fine-tuning phase, the model retains these state prediction and planning capabilities, thereby enabling it to perform implicit subtask planning even without explicitly utilizing interleaved planning during inference.

\subsubsection{\textbf{Effectiveness of Visual and Language Modalities in \textbf{Interleaved Planning}}}
To verify the effectiveness of the interleaved planning mechanism, we investigate the performance impact of omitting textual planning and keyframe forecasting, respectively.
\begin{itemize}
    \item Linguistic Planning: The results in Table~\ref{tab:sim_results} demonstrate that employing textual planning with BagelVLA in RoboTwin environment improves the success rate by 21\%. Similarly, in the two categories of real-world long-horizon tasks shown in Table~\ref{tab:long_horizon_tasks}, the use of textual planning also yields substantial performance gains. These two sets of experiments conclusively prove that incorporating language planning significantly benefits long-horizon tasks.
    
    \item Visual Forecasting: The keyframe variant in Table~\ref{tab:sim_results} and \ref{tab:long_horizon_tasks} illustrates the impact of using keyframe prediction as a training objective. The results indicate that visual planning markedly improves the accuracy of action planning in both simulation and real-world environments.
\end{itemize}
The aforementioned experiments confirm that both visual planning and language planning play crucial roles within the interleaved planning framework.

\section{Conclusion} 
\label{sec:conclusion}
We presented BagelVLA, a unified Vision-Language-Action framework 
for long-horizon manipulation by interleaving linguistic planning, visual forecasting, and action generation within a single transformer system. Building on Bagel’s unified multimodal backbone, we introduce an action expert and adopt a two-stage training recipe to progressively transfer multimodal reasoning and visual dynamics into embodied planning, then couple these representations with control. To address the latency of visual foresight, we further propose Residual Flow Guidance (RFG), which captures task-relevant future dynamics with substantially reduced computational costs. Overall, our results suggest that explicitly coupling linguistic planning with predictive visual representations can improve robustness and instruction-following in long-horizon manipulation.

\clearpage

\section{Acknowledgements}
We sincerely thank Weiwei Fang, Ziyang Liu, Zhelun Shi, Haitong Wang and Tingshuai Yan for their strong support and fruitful discussions.

\bibliographystyle{plainnat}
\bibliography{main}

\clearpage

\beginappendix

\section{Details of Model Architecture}
\label{sec:app-model}
The architecture of each expert in BagelVLA is detailed in the table below.
\begin{table}[h]
\caption{Parameters of Model Architecture
} 
\label{tab:model}      
\centering
\scriptsize
\begin{adjustbox}{width=\textwidth}
\begin{tabular}{c|c|c|c}
\toprule
\multicolumn{1}{c}{\textbf{Modules}}&\multicolumn{1}{c}{\textbf{Understanding Expert}}&\multicolumn{1}{c}{\textbf{Generation Expert}} & \multicolumn{1}{c}{\textbf{Action Expert}}\\
\midrule
Size & 7B & 7B&2B \\
Input Modality &Image/Text&Image&Proprio/Action\\
Output Modality & Text&Image&Action\\
Encoder & ViT+MLP &VAE+MLP&MLP\\
Image Resolution &256x256&256x256 (VAE)&-\\
Hidden size & 3584&3584&3584\\
Intermediate size & 18944 & 18944&3584 \\
Layers & 28 & 28&28 \\
Loss Type&CE&MSE(FM)&MSE(FM)\\
FM Timestep Distribution &-&LogitNormal(0, 1)&Beta(1.5,1)\\

\bottomrule
\end{tabular}
\end{adjustbox}
\vspace{-2ex}
\end{table}

\section{Dual Denoise Flow-Matching Implementation Details}
\label{sec:app-dualfm}
\begin{figure}[h]
    \centering
    \includegraphics[width=\textwidth]{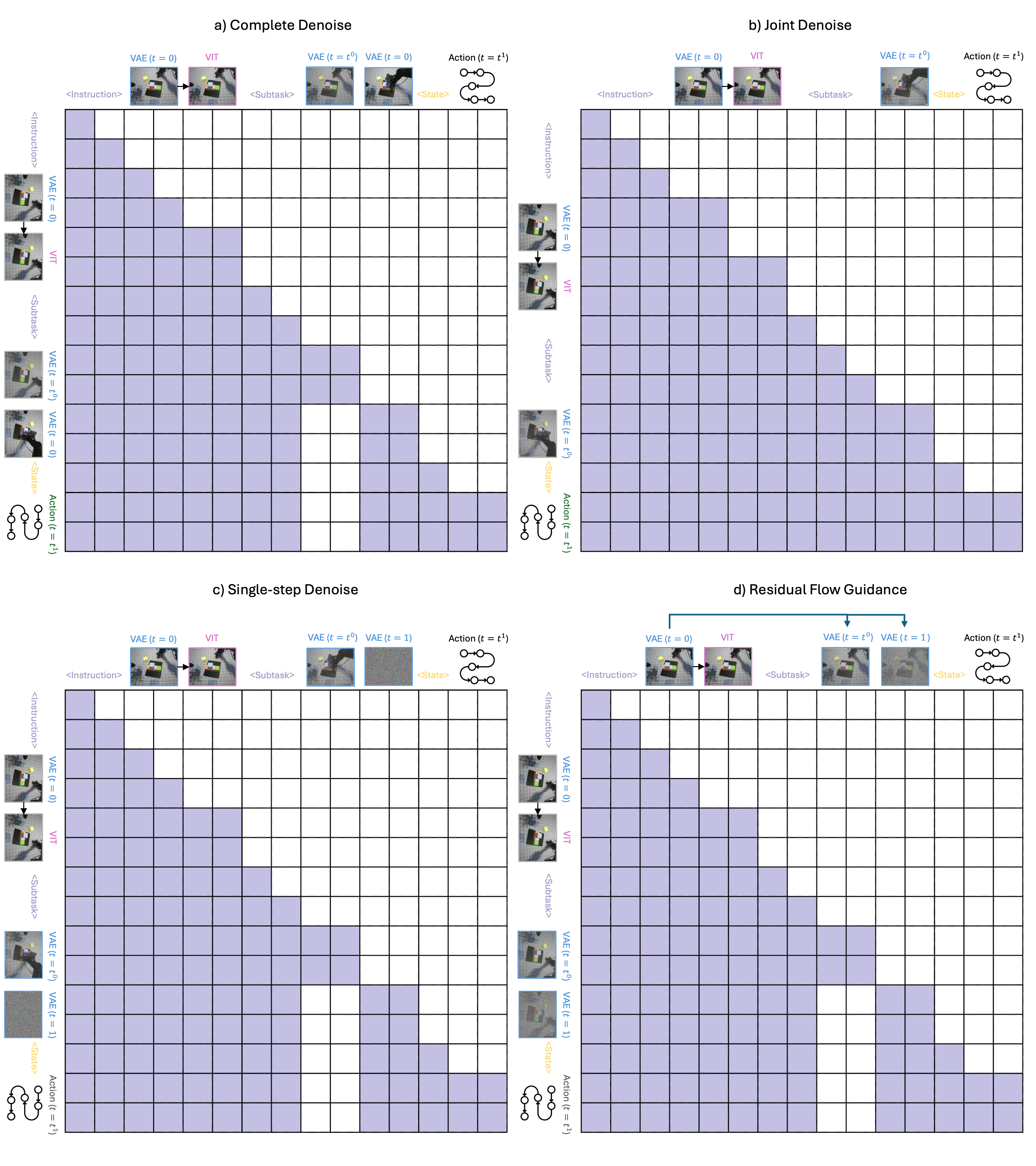}
    \caption{\textbf{Attention Mask used for Different Conditioning Schemes.} (a) Complete Denoise: Image prediction and action generation are performed separately, requiring a total of $N_1+N_2$ denoising steps.
(b) Joint Denoise: Image prediction and action generation are performed simultaneously, denoising together for $N$ steps.
(c) Single-Step Denoise: Action generation is conditioned directly on the context from the first denoising step of the image prediction. Further implementation details, including the construction of the concatenated sequence and the attention mask are provided in Appendix \ref{sec:app-dualfm}.
} 
    \label{fig:app-dualfm}
\end{figure}

Here, we demonstrate how to implement the three dual flow-matching methods mentioned in Sec.~\ref{sec:dualfm}. Specifically, this requires designing a unified multi-task attention mask for training, enabling a single input sequence to be used for the simultaneous computation of multiple task losses. When designing the corresponding interleaved sequences, we must not only prevent information leakage between different modalities but also align the training setup with special conditions encountered during inference, such as the time-sampling discrepancies that arise from varying numbers of denoising steps. We visualize the masking strategy used in our experiments in Fig.~\ref{fig:app-dualfm}.

\section{Data Details}
\label{sec:app-data}
\subsection{Stage 1: Pretraining - Finetuning Language Planning and Learning Visual Dynamics}
In this stage, we exclusively finetune the Understanding and Generation Experts to acquire capabilities in sub-task planning and keyframe prediction. To preserve the model's general linguistic proficiency, we co-train with general Question-Answering (QA) data. Specifically, the pretraining dataset comprises:
\begin{itemize}
    \item General VQA (Language Co-training): 2.56M QA pairs.
    \item Human-hand Data (Visual Dynamics): 310k episodes.
    \item Open-source Robot Data (Language Planning \& Visual Dynamics): 382k episodes.
    \item Self-collected Real Robot Data (Language Planning \& Visual Dynamics): 4.5k episodes.
\end{itemize}

\begin{table}[ht]
\caption{Details of Data component.}
\label{tab:ti2e-data}
\vspace{2mm}
\centering
\begin{tabular}{lccc}
\toprule
\textbf{Task name} &\textbf{Dataset name} &\textbf{Number of samples} \\
\midrule
\textbf{General VQA} & LLaVA-Pretrain\citep{liu2023visualinstructiontuning} & 558k \\
& FineVision\citep{wiedmann2025finevisionopendataneed} & 2M \\
\midrule
& AgibotWorld\citep{agibotworldcontributors2025agibotworldcolosseolargescale} & 120k \\
& GR\citep{cheang2025gr3technicalreport} & 80k \\
\textbf{Open-source Robot Data} & Galaxea Open-World\citep{jiang2025galaxeaopenworlddatasetg0} & 99k \\
& Bridge\citep{walke2023bridgedata} &55k \\
& Robotwin\citep{walke2023bridgedata} &27.5k \\
\midrule
\textbf{Human-hand Data}& Egodex\citep{hoque2025egodex} & 310k \\
\midrule
\textbf{Self-collected Data}& Aloha & 4.5k \\
\bottomrule
\end{tabular}
\end{table}

\subsection{Stage 2: Finetuning - Learning Action Planning}
In this stage, we introduce downstream robot data containing action labels for finetuning. We finetune the entire model on all three planning tasks simultaneously to obtain an interleaved planning model that performs robustly in specific scenarios. For the four scenarios used in our experiments, we employ the following finetuning strategies:
\begin{itemize}
    \item \textbf{Calvin (Visual Dynamics \& Action Planning)}: ABC dataset.
    \item \textbf{Robotwin (Language Planning, Visual Dynamics \& Action Planning)}: 50 tasks with 50 episodes each, totaling 2.5k episodes.
    \item \textbf{Aloha Short-horizon Tasks (Visual Dynamics \& Action Planning)}: 3k episodes.
    \item \textbf{Aloha Long-horizon Tasks (Visual Dynamics \& Action Planning)}: 1.5k episodes.
\end{itemize}

\subsection{Implementation Details about task Annotation}
\begin{figure}[h]
    \centering
    \includegraphics[width=0.8\textwidth]{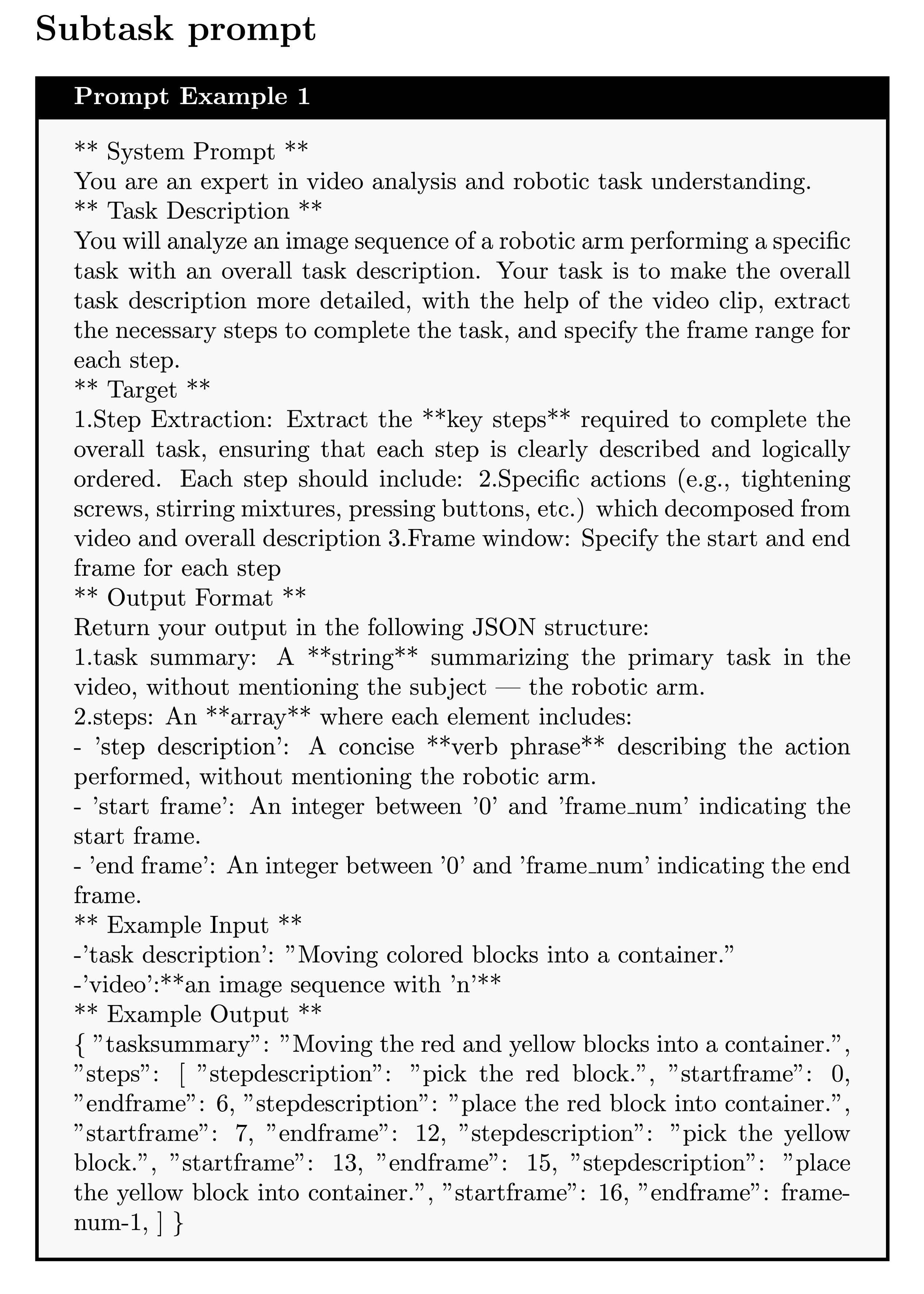}
    \caption{Prompt for Seed-1.5-VL-thinking.}
    \label{fig:prompt1}
\end{figure}

\begin{figure}[h]
    \centering
    \includegraphics[width=0.80\textwidth]{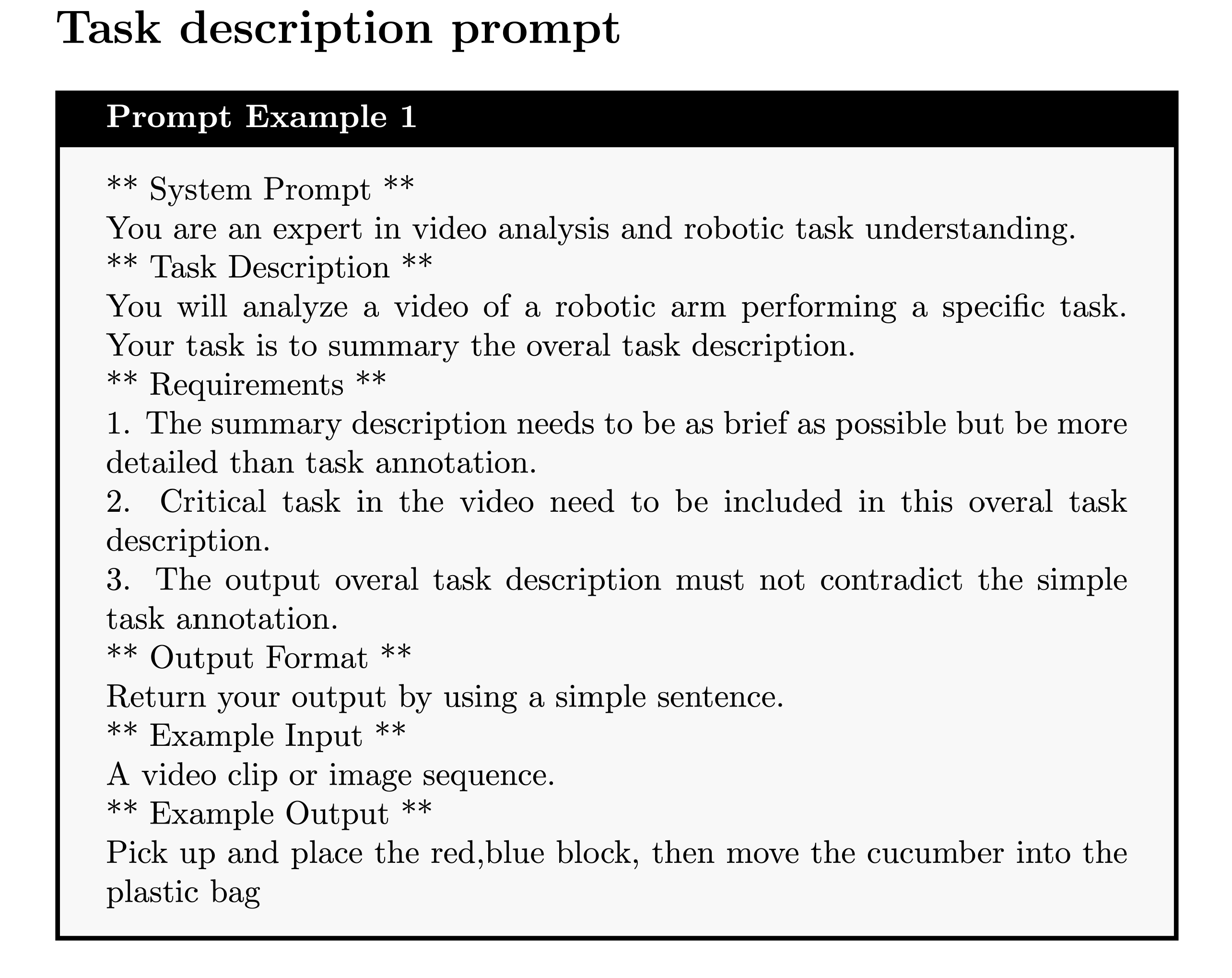}
    \caption{Prompt for Seed-1.5-VL-thinking.}
    \label{fig:prompt2}
\end{figure}

For open-source robotic datasets without subtask annotations, such as Bridge, we use the prompt template in Fig.~\ref{fig:prompt1}  and apply Seed-1.5-VL-thinking to process videos (or image sequences) solely for pretraining. For datasets that do not provide the overal task descriptions(e.g., EgoDex, AgiBot), we adopt the prompt template in Fig.~\ref{fig:prompt2} to extract a global task description used for planning or keyframe prediction.

\section{Training and Evaluation Details}
\label{sec:app-impl}
For all our experiments, we used a learning rate of 1e-5 and employed packed datasets within the FSDP framework to maximize resource utilization. Pre-training was conducted on 64 A800 GPUs with a batch size of approximately 1600 for 20,000 steps. For action fine-tuning and evaluation, we adopted different settings for various downstream scenarios:
\begin{itemize}
    \item Calvin ABC-D Simulation Environment: We trained on 8 A800 GPUs (effective batch size 192) for 30,000 steps. We used an action chunk size of 10, did not include proprioceptive input, and used two camera views as input to predict only the third view. For evaluation, we tested on 1,000 tasks of length 5 from the D-split and reported the mean task completion length.
    \item Robotwin Simulation Environment: We trained on 8 A800 GPUs using 2,500 clean demonstrations (effective batch size 128) for 60,000 steps. We used an action chunk size of 16, sampling one action every 3 steps (effective action horizon of 48). All three camera views were used as input, and we predicted the primary view image. For evaluation, we tested 100 times on 50 tasks in both Clean and Randomized settings using unseen instructions and reported the success rate.
    \item Real-Robot Tasks: We trained on 32 A800 GPUs (effective batch size 512) for 50,000 steps. We used an action chunk size of 24, inputting three views (primary, left wrist, right wrist) and predicting the primary view image. For evaluation, we tested each task type 20 times with randomized initial positions and added distractor objects. For OOD tasks, we included unseen target objects.
\end{itemize}

\section{Detailed Results in Simulation Environments}
\label{sec:app-sim_details}
Table \ref{tab:app-calvin} and \ref{tab:app-robotwin} present more detailed experimental results from the simulation environments.
\begin{table*}[h]
    \centering
    \caption{Evaluation on RoboTwin 2.0 Simulation (Clean vs Randomized, 50 tasks). The table shows success rates in percent (\%) for various models. Models are trained using 50 clean demos per task, and evaluated using unseen instructions.}
    \label{tab:app-robotwin}
    \resizebox{\textwidth}{!}{
    \begin{tabular}{c|rr|rr|rr|rr|rr|rr} 
        \toprule
        \multirow{2}{*}[-0.5ex]{\textbf{Robotwin Tasks}} & \multicolumn{2}{c}{\textbf{$\pi_0$}} & \multicolumn{2}{c}{\textbf{RDT}} & \multicolumn{2}{c}{\textbf{UP-VLA}}  & \multicolumn{2}{c}{\textbf{w/o Textual}} & \multicolumn{2}{c}{\textbf{w/o Keyframe}} & \multicolumn{2}{c}{\textbf{BagelVLA}} \\ 
        \cmidrule(lr){2-3} \cmidrule(lr){4-5} \cmidrule(lr){6-7} \cmidrule(lr){8-9} \cmidrule(lr){10-11} \cmidrule(lr){12-13} 
        & Clean & Random & Clean & Random & Clean & Random & Clean & Random & Clean & Random & Clean & Random \\ 
        \midrule
        Adjust Bottle & 90 & 56 & 81 & \textbf{75} & \textbf{100} & 17 & \textbf{100} & 7 & 99 & 4 & \textbf{100} & 14 \\
        Beat Block Hammer & 43 & 21 & 77 & \textbf{37} & 66 & 16 & 63 & 18 & 80 & 13 & \textbf{87} & 16 \\
        Blocks Ranking Rgb & 19 & 5 & 3 & 0 & 38 & 0 & 32 & 2 & 46 & \textbf{25} & \textbf{84} & 4 \\
        Blocks Ranking Size & 7 & 1 & 0 & 0 & 21 & 0 & 19 & 0 & 23 & \textbf{5} & \textbf{45} & 2 \\
        Click Alarmclock & 63 & 11 & 61 & 12 & 69 & 41 & 84 & \textbf{60} & 95 & 43 & \textbf{85} & 20 \\
        Click Bell & 44 & 3 & 80 & 9 & 54 & \textbf{72} & 78 & 60 & 98 & 29 & \textbf{100} & 35 \\
        Dump Bin Bigbin & 83 & 24 & 64 & 32 & 81 & 35 & 67 & 26 & 87 & 41 & \textbf{91} & \textbf{51} \\
        Grab Roller & 96 & \textbf{80} & 74 & 43 & 99 & 28 & \textbf{100} & 63 & 97 & 37 & 99 & 41 \\
        Handover Block & \textbf{45} & 8 & \textbf{45} & \textbf{14} & 4 & 0 & 0 & 0 & 18 & 1 & 38 & 0 \\
        Handover Mic & \textbf{98} & 13 & 90 & \textbf{31} & 45 & 0 & 76 & 0 & 44 & 3 & 75 & 8 \\
        Hanging Mug & 11 & 3 & \textbf{23} & \textbf{16} & 4 & 0 & 6 & 0 & 2 & 1 & 12 & 1 \\
        Lift Pot & 84 & \textbf{36} & 72 & 9 & 20 & 0 & 0 & 0 & 64 & 7 & \textbf{87} & 32 \\
        Move Can Pot & 58 & \textbf{21} & 25 & 12 & 48 & 0 & 51 & 0 & 9 & 2 & \textbf{78} & 0 \\
        Move Pillbottle Pad & 21 & 1 & 8 & 0 & 51 & \textbf{7} & 60 & 2 & 22 & 3 & \textbf{92} & 1 \\
        Move Playingcard Away & 53 & 22 & 43 & 11 & 79 & 13 & 86 & 6 & 64 & \textbf{31} & \textbf{92} & 30 \\
        Move Stapler Pad & 0 & \textbf{2} & 2 & 0 & 8 & 0 & 5 & 0 & 6 & 1 & \textbf{27} & 0 \\
        Open Laptop & 85 & \textbf{46} & 59 & 32 & 86 & 21 & 57 & 13 & 62 & 3 & \textbf{96} & 37 \\
        Open Microwave & \textbf{80} & \textbf{50} & 37 & 20 & 2 & 7 & 0 & 5 & 8 & 14 & 0 & 0 \\
        Pick Diverse Bottles & 27 & 6 & 2 & 0 & 52 & 18 & 74 & 22 & 15 & 11 & \textbf{83} & \textbf{34} \\
        Pick Dual Bottles & 57 & 12 & 42 & 13 & 82 & 31 & 89 & 33 & 33 & 9 & \textbf{93} & \textbf{56} \\
        Place A2b Left & 31 & 1 & 3 & 1 & 74 & 4 & 59 & 7 & 50 & \textbf{15} & \textbf{79} & 12 \\
        Place A2b Right & 27 & 6 & 1 & 1 & 56 & 1 & 53 & 6 & 55 & \textbf{19} & \textbf{81} & 11  \\
        Place Bread Basket & 17 & 4 & 10 & 2 & 63 & 20 & 71 & \textbf{29} & 42 & 17 & \textbf{90} & \textbf{29} \\
        Place Bread Skillet & 23 & 1 & 5 & 1 & 71 & 16 & 82 & \textbf{26} & 62 & 2 & \textbf{91} & \textbf{26} \\
        Place Burger Fries & 80 & 4 & 50 & 27 & 97 & 26 & 95 & \textbf{56} & 55 & 2 & \textbf{99} & 11 \\
        Place Can Basket & 41 & \textbf{6} & 19 & \textbf{6} & 20 & 0 & 37 & 1 & 8 & 0 & \textbf{63} & 0 \\
        Place Cans Plasticbox & 34 & 2 & 6 & 5 & 66 & 24 & 23 & \textbf{40} & 46 & 6 & \textbf{94} & 5 \\
        Place Container Plate & 88 & 45 & 78 & 17 & 86 & 48 & 97 & \textbf{71} & 82 & 55 & \textbf{100} & 58 \\
        Place Dual Shoes & 15 & 0 & 4 & 4 & 45 & 0 & 36 & \textbf{12} & 21 & 0 & \textbf{57} & 0 \\
        Place Empty Cup & 37 & 11 & 56 & 7 & 74 & 27 & 94 & \textbf{34} & 76 & 35 & \textbf{97} & \textbf{34} \\
        Place Fan & 20 & \textbf{10} & 12 & 2 & 31 & 1 & 15 & 3 & 18 & 2 & \textbf{62} & 5 \\
        Place Mouse Pad & 7 & 1 & 1 & 0 & 27 & 0 & 14 & 10 & 18 & 12 & \textbf{46} & \textbf{14} \\
        Place Object Basket & 16 & 2 & 33 & \textbf{17} & 56 & 1 & 44 & 1 & 40 & 6 & \textbf{66} & 3 \\
        Place Object Scale & 10 & 0 & 1 & 0 & 36 &  4 & 46 & 7 & 31 & \textbf{8} & \textbf{71} & 0 \\
        Place Object Stand & 36 & 11 & 15 & 5 & 76 & 24 & 77 & \textbf{35} & 45 & 27 & \textbf{87} & 21 \\
        Place Phone Stand & 35 & \textbf{7} & 15 & 6 & 32 & 0 & 48 & 0 & 33 & 9 & \textbf{61} & 2 \\
        Place Shoe & 28 & 6 & 35 & 7 & 76 & 12 & 63 & 15 & 44 & 23 & \textbf{90} & \textbf{29} \\
        Press Stapler & 62 & 29 & 41 & 24 & 79 & 56 & 59 & 50 & 93 & 52 & \textbf{94} & \textbf{58} \\
        Put Bottles Dustbin & \textbf{54} & \textbf{13} & 21 & 4 & 7 & 0 & 12 & 0 & 10 & 0 & 42 & 10 \\
        Put Object Cabinet & \textbf{68} & \textbf{18} & 33 & \textbf{18} & 7 & 0 & 45 & 4 & 21 & 1 & 52 & 0 \\
        Rotate Qrcode & 68 & 15 & 50 & 5 & 56 & 2 & 68 & 3 & 72 & 4 & \textbf{81} & \textbf{21} \\
        Scan Object & 18 & 1 & 4 & 1 & 47 & 23 & 66 & 22 & 38 & 3 & \textbf{77} & \textbf{32} \\
        Shake Bottle Horizontally & 99 & 51 & 84 & 51 & \textbf{100} & 68 & 99 & \textbf{84} & 87 & 60 & \textbf{100} & 73 \\
        Shake Bottle & 97 & 60 & 74 & 45 & 98 & 54 & 98 & \textbf{82} & 83 & 44 & \textbf{100} & 74 \\
        Stack Blocks Three & 17 & 0 & 2 & 0 & 8 & 0 & 15 & 0 & 5 & 2 & \textbf{45} & \textbf{5} \\
        Stack Blocks Two & 42 & 1 & 21 & 2 & 61 & 0 & 59 & 2 & 29 & \textbf{31} & \textbf{95} & 6 \\
        Stack Bowls Three & 66 & \textbf{24} & 51 & 17 & 42 & 1 & 42 & 7 & 37 & 12 & \textbf{63} & 13 \\
        Stack Bowls Two & \textbf{91} & 41 & 76 & 30 & 69 & 12 & 70 & 21 & 88 & 48 & 90 & \textbf{52} \\
        Stamp Seal & 3 & 4 & 1 & 0 & 34 & 2 & 29 & 1 & 23 & \textbf{8} & \textbf{77} &  \textbf{8}\\
        Turn Switch & 27 & 23 & 35 & 15 & 43 & 26 & 37 & 14 & \textbf{52} & 10 & 49 & \textbf{30} \\
        \midrule
        \textbf{Average} & 46.42 & 16.34 & 34.50 & 13.72 & 52.92 & 15.16 & 54.00 & 19.20 & 56.72 & 15.92 & \textbf{75.26} & \textbf{20.87} \\
        \bottomrule
    \end{tabular}
    }
\end{table*}
\begin{table*}[h]
    \caption{Detailed results of evaluation on the Calvin ABC$\rightarrow$D benchmark.}
\label{tab:app-calvin}
    \centering
    \scriptsize
    \begin{adjustbox}{width=0.65\linewidth}
    \begin{tabular}{lrrrrrr}
    \toprule
    \multirow{2}{*}[-0.5ex]{\textbf{Method}} & \multicolumn{5}{c}{\textbf{Tasks completed in a row}} & \multirow{2}{*}[-0.5ex]{\textbf{Avg. Len $\uparrow$}} \\ 
    \cmidrule(lr){2-6} 
     & \textbf{1} & \textbf{2} & \textbf{3} & \textbf{4} & \textbf{5} & \\ \midrule
    $\pi_0$* & 0.937 & 0.832 & 0.740& 0.629 &0.510 & 3.65\\ 
    UP-VLA & 0.928 &  0.865 &  0.815 &  0.769 &  0.699 &  4.08 \\
    VPP & 0.965 & 0.909 & 0.866 & 0.820 &0.769 & 4.33\\ 
    w/o Keyframe-forecasting&0.909 & 0.792 & 0.676 & 0.546 &0.422 & 3.35\\
    \rowcolor{skyblue} BagelVLA (Ours) &\textbf{0.993} & \textbf{0.954}& \textbf{0.893 }&\textbf{ 0.824}&\textbf{0.741} & \textbf{4.41}\\ 
    \bottomrule
    \end{tabular}
    \end{adjustbox}
\end{table*}

\section{Evaluation Demos of Real-World Tasks}
\label{sec:app-demo}
In this section, we detail the setup for two categories of real-robot tasks: Basic Tasks and Long-Horizon Planning Tasks. We also present demo videos of BagelVLA performing on each task type.
\subsection{Basic Tasks}
\label{sec:app-basic}
During testing, we incorporate several kinds of randomness to evaluate robustness and generalization: \textit{Novel Objects}: Adding unseen objects. \textit{Distractors:} Operating in the presence of irrelevant distractor objects. \textit{Visual Variations:} Adapting to changes in background color and object color.
The task suite for the basic tasks on the 14-DOF dual arm includes:
\begin{itemize}
    \item \textbf{Pick \& Place:} Grasping and placing a wide range of objects. The training set includes toy fruits, a computer mouse, colorful blocks, toy phones, and so on. The placed targets include colorful plates, baskets, boxes, and so on.
        \item \textbf{Pick \& Place Unseen:} Grasping and placing unseen objects to unseen targets. We tested picking up OOD objects such as pears, peaches, a purple block, and placing to novel targets, like pink plates, transparent plates, pink blocks, and so on. We found that although the training set scenes did not involve numerous distractor objects or unseen items, the model still robustly generalizes to new objects and targets with the correct semantics.
    
    \item \textbf{Water Flower:} This task involves grasping the handle of a toy watering can to simulate the pouring action of watering a plant. It rigorously tests the model's fine-grained manipulation capabilities, as any action error could easily result in a failure to grasp the handle or align with the flowerpot.
    
    \item \textbf{Stack Cubes:} The training data includes blocks of four different colors. The instructions require stacking several of these blocks together (up to three high), but without a specific order.
    
    \item \textbf{Put Flowers in Vase:} Grasp a bouquet lying flat on the table and insert it into a vase. This task requires the model to precisely grasp the thin stems of the bouquet and align them with the opening of the vase, testing the accuracy of the manipulation.

    \item \textbf{Stack Bowls:} Stack bowls of three different colors according to a specified color sequence. This task evaluates the model's robustness to object positions and its ability to follow language instructions.

    \item \textbf{Pour Fries:} Open the lid of a carton and pour the toy fries from inside it onto a plate. This is a relatively long-horizon task that requires the model to autonomously determine the next action based on its current progress. It tests both manipulation accuracy and long-horizon task capabilities.

    \item \textbf{Sweep Rubbish:} Grasp a toy broom, sweep the randomly placed tissue paper trash on the table into a dustpan, and then put down the broom. This is a task that combines both long-horizon planning and dynamic control. The model must not only assess its current progress but also increase the sweeping speed to ensure the tissue paper rolls into the dustpan.

    \item \textbf{Press Button:} Press different buttons in a specified color sequence. This is a simple long-horizon task that also tests the model's semantic following capabilities.
    
    \item \textbf{Drawer Operation:} Opening and closing a drawer. This task primarily evaluates the accuracy of the manipulation.
\end{itemize}

\begin{figure}[h]
    \centering
    \includegraphics[width=\textwidth]{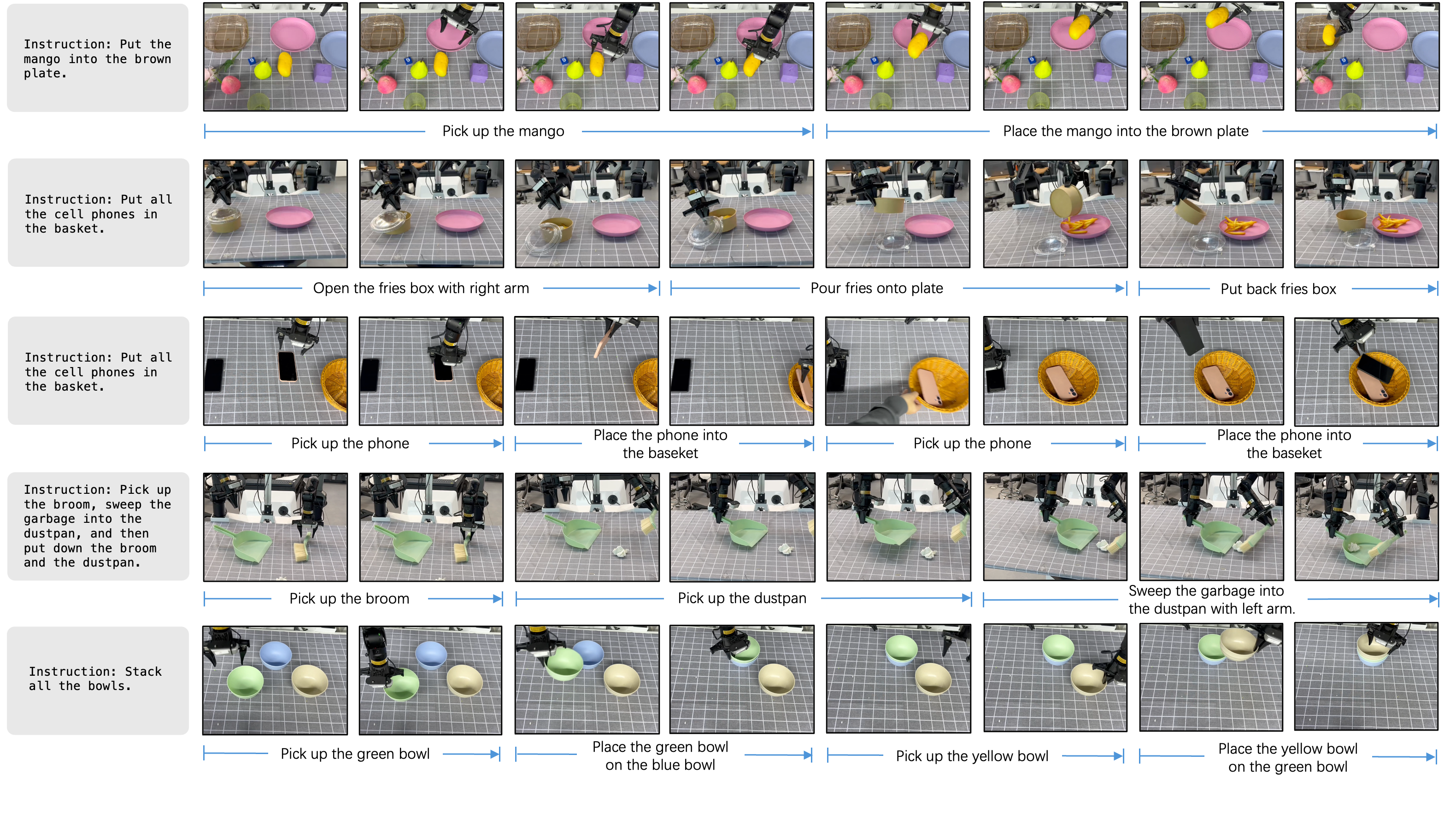}
    \caption{\textbf{Demos videos of BagelVLA on Basic Tasks}.
} 
    \label{fig:app-basicdemo}
\end{figure}

Fig.~\ref{fig:app-basicdemo} illustrates several test scenarios for the basic tasks and presents video recordings of the model's performance.

\subsection{Long-Horizon Planning Tasks}
\label{sec:app-hard}
We designed two distinct types of long-horizon planning tasks: (1) Stack Cubes in Requested Order and (2) Calculate and Place Symbol Blocks. We will now detail the setup for each and showcase corresponding demonstration videos.

\paragraph{\textbf{Stack Cubes in Requested Order}}
This task requires the model to stack scattered, multi-colored cubes from the tabletop into a structure that matches a specified shape and sequence given by a language instruction. The target structures can range from one to three layers, with each layer containing one to three cubes. An example instruction is: \textit{Place the cubes in order: the first layer is a blue and a green block, the second layer is an orange block.} The model must perform interleaved planning at each step based on this high-level command. This task involves a very long sequence of actions, posing a significant semantic-following challenge for conventional methods that do not employ explicit planning. In our experiments in Sec.~\ref{sec:exp-hard}, we demonstrate that our method holds a distinct advantage on such long-horizon tasks.

\paragraph{\textbf{Calculate and Place Symbol Blocks}}
This task requires the model to assemble scattered number and symbol blocks to form an arithmetic equation specified by a language instruction, such as: \textit{Assemble the building blocks to complete the equation 21+3=?} The initial scene may already contain partially arranged blocks, forcing the model to autonomously decide which block to grasp and place next. It must also place the correct blocks representing the calculated result. Similar to the stacking task, this task also involves long-horizon operational planning. Beyond that, it introduces an additional layer of complexity by requiring a Chain-of-Thought (CoT) process: the model must first leverage the mathematical reasoning capabilities of the general-purpose VLM to compute the result, and then map this result back to the planning and action space. We use this task to validate the effectiveness and generalization capabilities of our interleaved planning framework on long-horizon reasoning tasks.

\begin{figure}[h]
    \centering
    \includegraphics[width=\textwidth]{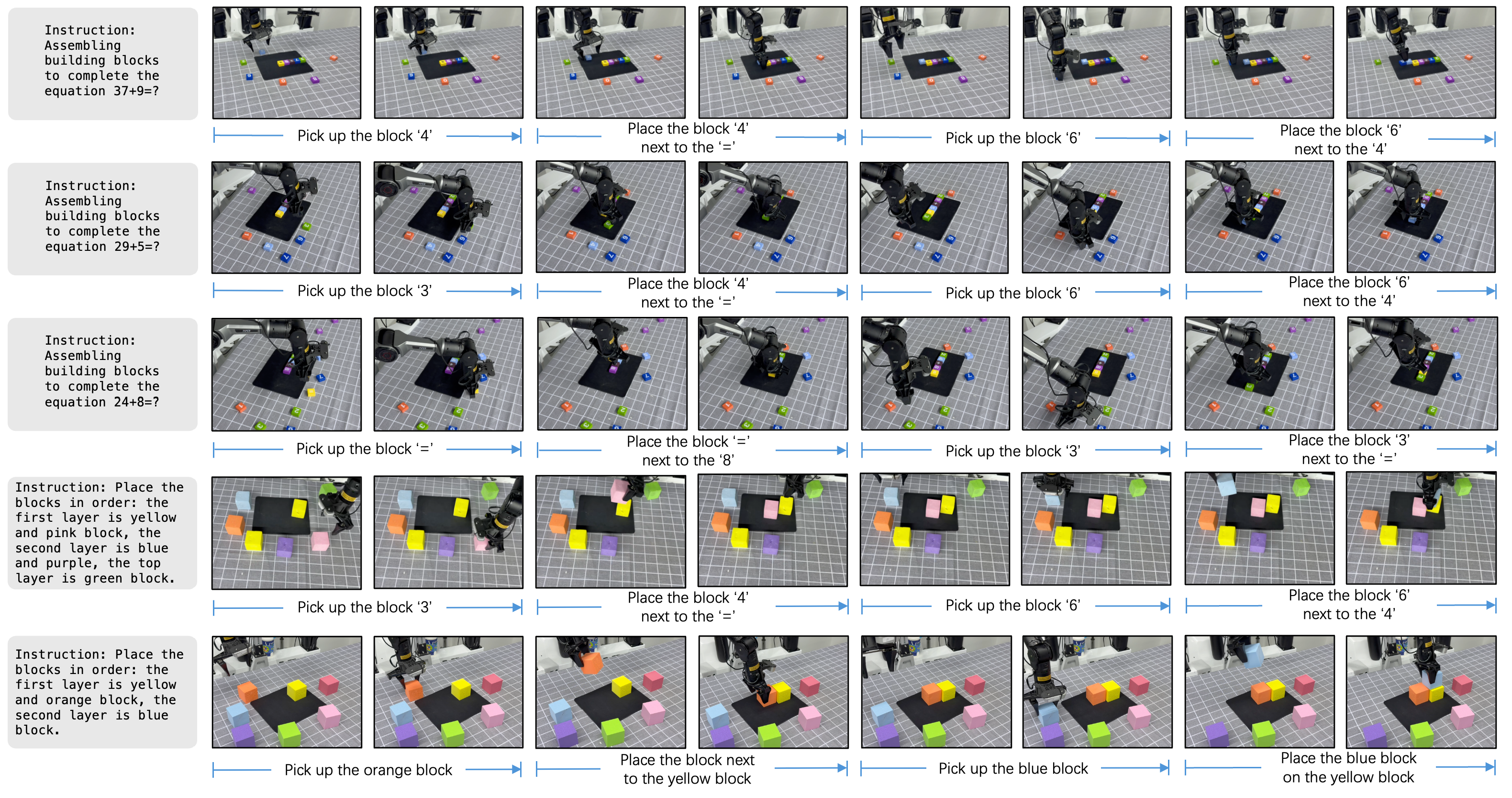}
    \caption{\textbf{Demos videos of BagelVLA on Long-Horizon Planning Tasks}.
} 
    \label{fig:app-lhdemo}
\end{figure}

Fig.~\ref{fig:app-lhdemo} illustrates several test scenarios for the long-horizon planning tasks and presents video recordings of the model's performance.

\section{More Interleaved Planning Visualizations on diverse robotic Tasks}
\label{sec:app-inter}
\begin{figure}[t]
    \centering
    \includegraphics[width=1.0\textwidth]{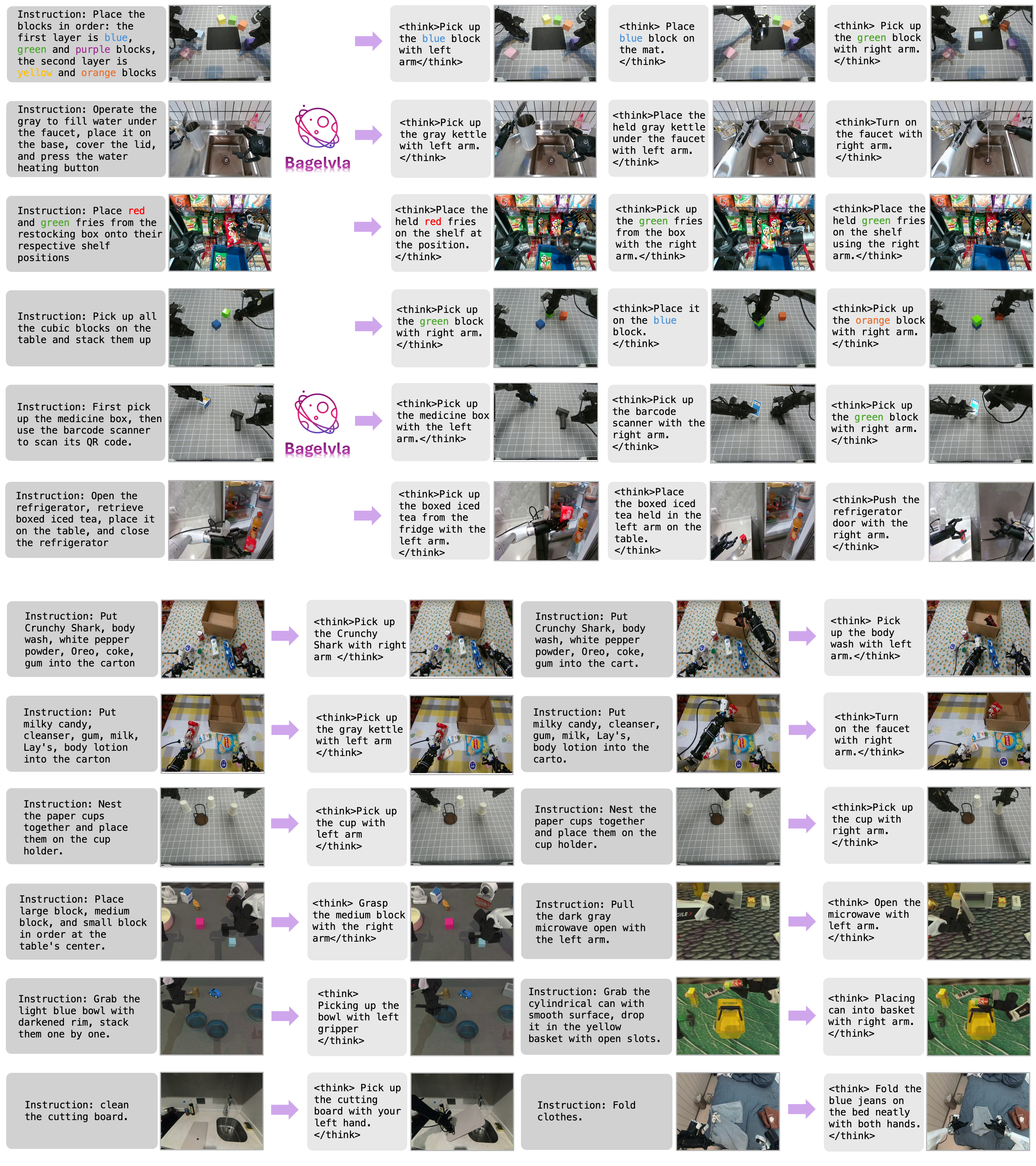}
    \caption{\textbf{Visualizations of interleaved planning results on diverse robotic tasks. 
    } Given a global instruction and the current observation, BagelVLA leverages the context to identify the immediate subtask, predicts a goal image for that subtask.}
    \label{fig:app-realwd}
\end{figure}

Similar to Fig.~\ref{fig:exp-realwd}, in Fig.~\ref{fig:app-realwd} we provide additional results of interleaved planning in real-world scenarios for reference.
\section{More Comparison using RFG and Naive Single-Step Denoising}
\label{sec:app-rfg}
\begin{figure*}[ht]
    \centering
    \includegraphics[width=1.0\textwidth]{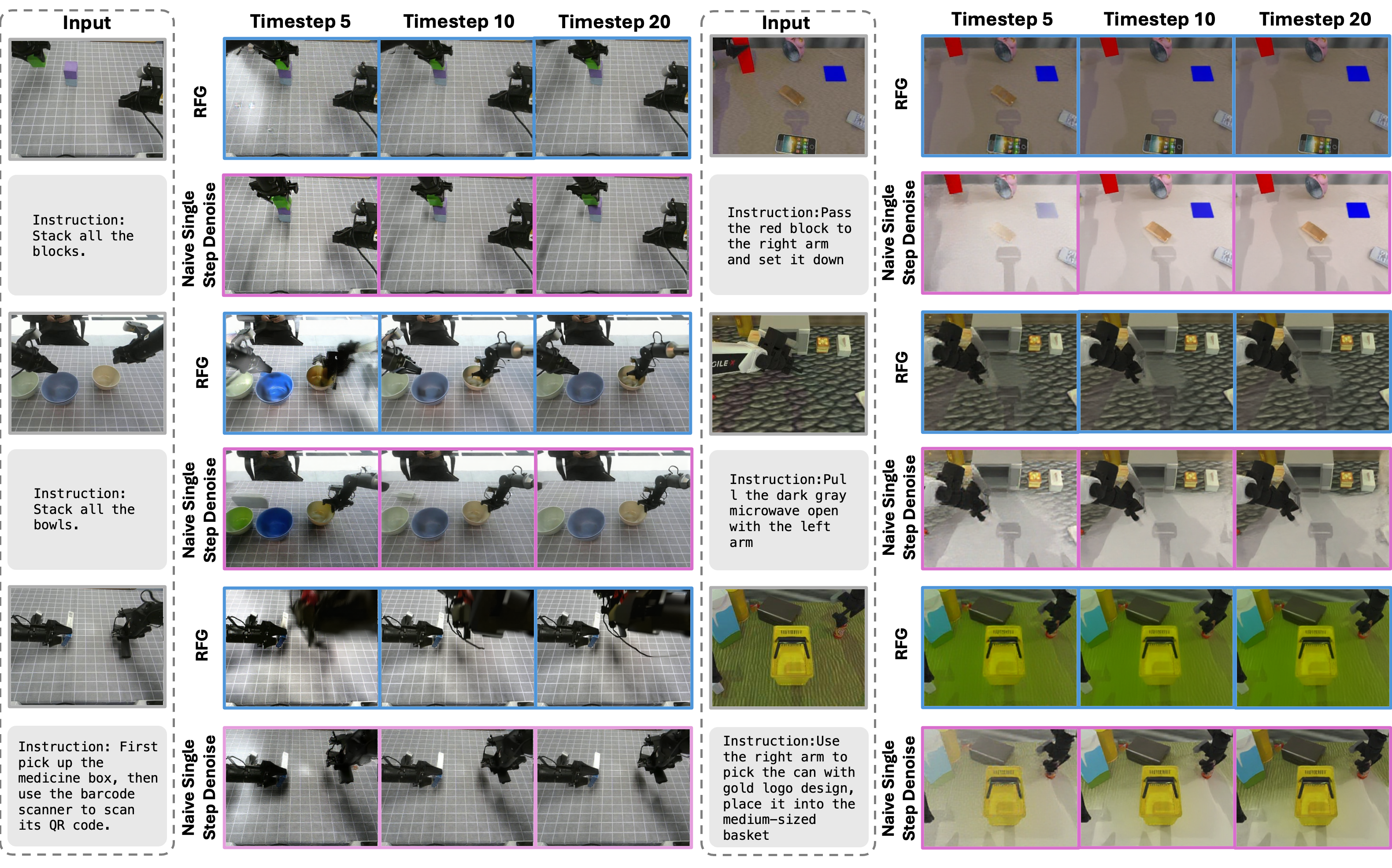}
    \caption{\textbf{Predicted images using different denoising steps. 
    } The figure displays the generation results for the naive single-step denoise (Eq.~\ref{vpp1}) and RFG (Eq.~\ref{vpp2}) across varying denoising steps in real-world tasks and simulation scenarios. RFG demonstrates the capability to preserve backgrounds and achieve high-quality generation with very few steps. This provides strong support for reducing the inference latency of interleaved generation.}
    \label{fig:app-keyframe}
\end{figure*}

In Fig.~\ref{fig:app-keyframe}, we provide additional comparison using RFG and naive single-step denoising for reference.

\section{Usage of LLMs}
In the final stages of preparing this manuscript, the authors used a Large Language Model (LLM) solely for grammar checking and language polishing. The model assisted in improving sentence structure and correcting grammatical errors to enhance readability. 

\end{document}